\documentclass[pdflatex,sn-mathphys-num]{sn-jnl}

\usepackage{graphicx}%
\usepackage{multirow}%
\usepackage{amsmath,amssymb,amsfonts}%
\usepackage{amsthm}%
\usepackage{mathrsfs}%
\usepackage[title]{appendix}%
\usepackage{xcolor}%
\usepackage{textcomp}%
\usepackage{manyfoot}%
\usepackage{booktabs}%
\usepackage{algorithm}%
\usepackage{algorithmicx}%
\usepackage{algpseudocode}%
\usepackage{listings}%
\usepackage{bbding}
\usepackage{array}

\usepackage{rotating}

\usepackage{bbm}
\usepackage{color,xcolor}
\usepackage{colortbl}
\definecolor{mygray}{gray}{.9}
\usepackage{caption}

\definecolor{myorange}{rgb}{0.9804, 0.6667, 0.1176}
\definecolor{mygreen}{rgb}{0, 0.8, 0}
\definecolor{myblue}{rgb}{0.92, 0.948, 1}
\definecolor{dark-red}{rgb}{0.6, 0.1, 0.1}
\definecolor{dark-green}{rgb}{0, 0.6, 0.25} 
\definecolor{citecolor}{rgb}{0,0.443,0.737} 
\definecolor{linkcolor}{rgb}{0.956,0.298,0.235} 

\newcommand{\rebut}[1]{{\color{black}#1}}

\theoremstyle{thmstyleone}%
\theoremstyle{thmstyletwo}%

\theoremstyle{thmstylethree}%

\begin{document}
	
	
	\title[Article Title]{UniBiomed: A Universal Foundation Model for Grounded Biomedical Image Interpretation}
	
	\author[1]{\fnm{Linshan} \sur{Wu}}
	
	\author[1]{\fnm{Yuxiang} \sur{Nie}}
	
	\author[1]{\fnm{Sunan} \sur{He}}
	
	\author[1]{\fnm{Jiaxin} \sur{Zhuang}}
	
	\author[1,2]{\fnm{Luyang} \sur{Luo}}
	
	\author[3]{\fnm{Tao} \sur{Li}}
	
	\author[3]{\fnm{Zhuoyao} \sur{Xie}}
	
	\author[3]{\fnm{Dexuan} \sur{Chen}}
	
	\author[3]{\fnm{Yinghua} \sur{Zhao}}
	
	\author[4]{\fnm{Neeraj} \sur{Mahboobani}}
	
	\author[5]{\fnm{Varut} \sur{Vardhanabhuti}}
	
	\author[6,7]{\fnm{Ronald Cheong Kin} \sur{Chan}}
	
	\author[8]{\fnm{Yifan} \sur{Peng}}
	
	\author[2]{\fnm{Pranav} \sur{Rajpurkar}}
	
	\author*[1,9,10,11,12]{\fnm{Hao} \sur{Chen}}\email{jhc@cse.ust.hk}
	
	\affil[1]{\orgdiv{Department of Computer Science and Engineering}, \orgname{The Hong Kong University of
			Science and Technology}, \orgaddress{\state{Hong Kong}, \country{China}}}
	
	\affil[2]{\orgdiv{Department of Biomedical Informatics}, \orgname{Harvard University}, \orgaddress{\state{Boston}, \country{USA}}}
	
	\affil[3]{\orgdiv{Department of Radiology}, \orgname{The
			Third Affiliated Hospital of Southern Medical University}, \orgaddress{\state{Guangzhou}, \country{China}}}
	
	\affil[4]{\orgdiv{Department of Imaging and Interventional Radiology}, \orgname{The Chinese University of Hong Kong}, \orgaddress{\state{Hong Kong}, \country{China}}}
	
	\affil[5]{\orgdiv{Department of Diagnostic Radiology}, \orgname{Li Ka Shing Faculty of Medicine, The University of Hong Kong}, \orgaddress{\state{Hong Kong}, \country{China}}}
	
	\affil[6]{\orgdiv{Department of Anatomical and Cellular Pathology}, \orgname{The Chinese University of Hong Kong}, \orgaddress{\state{Hong Kong}, \country{China}}}
	
	\affil[7]{\orgdiv{State Key Laboratory of Translational Oncology}, \orgname{The Chinese University of Hong Kong}, \orgaddress{\state{Hong Kong}, \country{China}}}
	
	\affil[8]{\orgdiv{Population Health Sciences}, \orgname{Weill Cornell Medicine}, \orgaddress{\state{New York}, \country{USA}}}
	
	\affil[9]{\orgdiv{Department of Chemical and Biological Engineering}, \orgname{The Hong Kong University of Science and Technology}, \orgaddress{\state{Hong Kong}, \country{China}}}
	
	\affil[10]{\orgdiv{Division of Life Science}, \orgname{The Hong Kong University of Science and Technology}, \orgaddress{\state{Hong Kong}, \country{China}}}
	
	\affil[11]{\orgdiv{State Key Laboratory of Molecular Neuroscience}, \orgname{The Hong Kong University of Science and Technology}, \orgaddress{\state{Hong Kong}, \country{China}}}
	
	\affil[12]{\orgdiv{Shenzhen-Hong Kong Collaborative Innovation Research Institute}, \orgname{The Hong Kong University of Science and Technology}, \orgaddress{\state{Shenzhen}, \country{China}}}

	\abstract{
		The integration of AI-assisted biomedical image analysis into clinical practice demands AI-generated findings that are not only accurate but also interpretable to clinicians.
		However, existing biomedical AI models generally lack the ability to simultaneously generate diagnostic findings and localize corresponding biomedical objects. 
		This limitation makes it challenging for clinicians to correlate AI-generated findings with visual evidence (\emph{e.g.}, tiny lesions) in images and interpret the results of AI models.
		To address this challenge, we introduce \textbf{UniBiomed}, the first universal foundation model for grounded biomedical image interpretation, which is capable of generating accurate diagnostic findings and simultaneously segmenting the corresponding biomedical targets. 
		UniBiomed is based on a novel integration of Multi-modal Large Language Model and Segment Anything Model, which can effectively unify diverse biomedical tasks in universal training for advancing grounded interpretation.   
		To develop UniBiomed, we curate a large-scale dataset comprising over 27 million triplets of images, region annotations, and text descriptions across ten biomedical imaging modalities. Extensive validation on 70 internal and 14 external datasets demonstrated the state-of-the-art performance of UniBiomed in diverse biomedical tasks, including image segmentation, disease recognition, region-aware diagnosis, vision question answering, and report generation.  
		In summary, UniBiomed is a powerful and versatile biomedical foundation model, unlocking the untapped grounded interpretation capability for optimizing AI-assisted biomedical image analysis.
	}
	
	
	\maketitle
	
	\section{Introduction}\label{sec1}
	
	Multi-modal interpretation of biomedical images opens up novel opportunities in biomedical image analysis~\cite{perez2025exploring,royer2023future,li2023challenges}. Visual information from biomedical imaging enables detailed anatomical and functional analysis from cell to organ levels~\cite{radsch2023labelling,engelmann2022detecting,saporta2022benchmarking,bilodeau2022microscopy,luo2020deep,xu2024whole,sun2024foundation,msd}, while the textual information from diagnostic findings provides fine-grained descriptions for interpreting imaging~\cite{peng2023ai,lu2024multimodal,schafer2024overcoming}. 
	However, it remains challenging for existing biomedical AI models to effectively integrate holistic vision and language information to assist clinicians in clinical practice.
	Although recent multi-modal biomedical foundation models~\cite{schafer2024overcoming,Biomedgpt,gsco,he2024meddr,kim2024transparent,Llava-med} have showcased encouraging results in interpreting biomedical images and generating diagnostic findings, these models are typically region-agnostic and fail to extract the target regions (\emph{e.g.}, tiny lesions) described in the generated findings. This limitation poses a critical barrier for clinicians to associate AI-derived findings with specific regions in images and interpret the results, significantly hindering the clinical practice of biomedical AI models.
	
	Biomedical image segmentation is a practical solution for extracting structured visual information from biomedical images~\cite{medsam,nnUNet,biomedparse,wang2021deep,archit2025segment,stringer2021cellpose,pachitariu2022cellpose,wu2025modeling,wu2023querying,wu2023sparsely,DBFNet,DCA,liu2023multi,ma2025generative,zhuang2025bio2vol,stringer2025cellpose3,peiris2023uncertainty}, enabling the identification of regions of interest (ROI) across organs, lesions, tissues, and cells. Although recent segmentation foundation models~\cite{medsam,biomedparse,segvol,SAT} have demonstrated remarkable performance in this task, they generally lack the ability to generate diagnostic findings (\emph{e.g.}, clinical reports), limiting their practicality in AI-assisted biomedical image analysis. 
	
	To address these challenges, we highlight the importance of grounded interpretation in biomedical image analysis, \emph{i.e.}, generating diagnostic findings and simultaneously segmenting the corresponding biomedical targets. 
	In this way, we enable biomedical AI models to deliver both accurate and interpretable results for effectively assisting clinicians in biomedical image analysis.
	In this work, we introduce \textbf{UniBiomed}, the first universal foundation model for grounded biomedical image interpretation. UniBiomed innovatively integrates advanced Multi-modal Large Language Models (MLLMs)~\cite{llava,internvl} and Segment Anything Model~\cite{sam,sam2} (SAM) for grounded biomedical image interpretation, as shown in \textbf{Figure~\ref{fig_overview}~(c)}. Specifically, the MLLM is used to interpret multi-modal biomedical images and generate diagnostic findings. Then, we combine the encoded user instructions and the outputs of MLLM to prompt the SAM model for segmenting the biomedical objects corresponding to the generated findings.
	Through this integration, UniBiomed can tackle a wide range of biomedical tasks, including biomedical image segmentation, disease recognition, region-aware diagnosis, vision question answering (VQA), and report generation, as shown in \textbf{Figure~\ref{fig_overview}~(d)}.
	These advancements enable UniBiomed to simultaneously provide fine-grained visual and textual information for fine-grained biomedical image analysis.
	
	The power of UniBiomed originates from its ability to leverage holistic multi-modal biomedical information in universal training. Concretely, previous biomedical models typically rely on disjoint training with independent datasets, \emph{e.g.}, clinical report datasets for report generation training~\cite{Biomedgpt,Llava-med,medtrinity,ctrate}, segmentation datasets for segmentation training~\cite{medsam,biomedparse,stringer2021cellpose}. In contrast, our UniBiomed is versatile in utilizing different types of biomedical datasets to improve the performance of biomedical image analysis complementarily. Within a universal training process, training with segmentation datasets enables UniBiomed to extract critical biomedical objects, region-aware diagnosis datasets enhance the region-aware ability, and VQA and report generation datasets improve the model's capability to interpret clinicians' instructions and produce accurate diagnostic findings. Integrating complementary multi-modal datasets for training, UniBiomed demonstrates more powerful and versatile abilities in biomedical image analysis.
	
	To develop UniBiomed, as shown in \textbf{Figure~\ref{fig_overview}~(a) and (b)}, we curate a large-scale dataset containing 27 million triplets of images, region annotations, and text descriptions spanning 10 biomedical imaging modalities. 
	Region annotations include both segmentation masks and bounding boxes, providing detailed spatial localization information for the model to develop region-aware capabilities. The text descriptions are extracted from readily available clinical texts accompanying public datasets, encompassing biomedical definitions, diagnostic findings, medical knowledge, and clinical reports. To the best of our knowledge, this is the largest and most comprehensive dataset for biomedical grounded interpretation. By integrating comprehensive and multi-granular visual and textual biomedical information for training, UniBiomed unleashes the untapped grounded interpretation capability in biomedical image analysis and achieves superior performance across diverse biomedical tasks.
	
	\begin{figure*}
		\centering
		\includegraphics[width=0.93\linewidth]{./fig/fig1.pdf}
	\end{figure*}
	\clearpage
	\begin{figure*}
		\centering
		\caption{\textbf{Overview of the study}. \textbf{a.} UniBiomed enables universal grounded interpretation across 10 different biomedical imaging modalities. CT, computed tomography; MRI, magnetic resonance imaging; OCT, optical coherence tomography; PET, positron emission tomography. \textbf{b.} We curated 27 million triplets of images, region annotations (segmentation masks and bounding boxes), and text descriptions for training UniBiomed. The text descriptions are processed from readily available diagnostic findings in public datasets. \textbf{c.} The framework of UniBiomed. UniBiomed incorporates Multi-modal Large Language Models (MLLMs)~\cite{llava,internvl} with Segment Anything Model~\cite{sam,sam2} (SAM) to tackle text description and segmentation tasks jointly. MLLM contains both vision and language models to interpret visual and language information, ultimately generating clinical text descriptions.
			SAM includes a vision encoder, a prompt encoder, and a mask decoder to address the segmentation task. Given user instructions with biomedical images, UniBiomed can segment target objects and generate grounded text descriptions simultaneously, enabling end-to-end analysis of multi-modal biomedical images. \textbf{d.} UniBiomed is designed to unify diverse biomedical tasks within a universal training process, including segmentation, disease recognition, region-aware diagnosis, vision question answering (VQA), and report generation.}
		\label{fig_overview}
	\end{figure*}
	
	We conduct a large-scale validation on 70 internal and 14 external datasets across 10 diverse biomedical imaging modalities. Extensive experiments demonstrate the effectiveness of UniBiomed. Specifically, in biomedical image segmentation, UniBiomed surpasses state-of-the-art segmentation foundation models~\cite{medsam,biomedparse,segvol,SAT} by a substantial margin, \emph{e.g.}, surpassing the representative model BiomedParse~\cite{biomedparse} by an average of $10.25\%$ in Dice scores across 60 segmentation datasets. 
	Beyond segmentation, UniBiomed also achieves superior performance in diverse biomedical tasks compared with state-of-the-art biomedical MLLMs, including disease recognition, VQA, ROI classification, region-aware report generation, and report generation. 
	More importantly, we further showcase that UniBiomed is an effective biomedical AI tool for optimizing the biomedical image analysis workflow. 
	Specifically, to recognize and localize abnormalities in images, previous biomedical AI models~\cite{medsam,biomedparse,segvol,SAT} heavily rely on clinical experts to provide accurate textual or visual prompts, \emph{e.g.}, instruct the model that the input image contains tumors (text prompts) or provide tight bounding boxes to highlight the tumors in the images (visual prompts).
	In contrast, UniBiomed can eliminate these processes and enable automated end-to-end grounded interpretation of biomedical images, effectively streamlining the analysis workflow. 
	In summary, UniBiomed represents a versatile and powerful foundation model that delivers superior performance in grounded biomedical image interpretation, demonstrating promising potential towards more accurate and efficient biomedical image analysis.
	
	\clearpage
	
	\section{Results}\label{sec2}
	
	\subsection{Overview of UniBiomed}
	\label{sec_result_Overview}
	
	\textbf{Dataset}. UniBiomed is designed to address the biomedical grounded interpretation tasks, necessitating training datasets to provide both region information and expert-annotated text descriptions. For the region information, we collect the readily available segmentation masks and bounding boxes from public biomedical datasets. Specifically, the segmentation masks are used for dense segmentation training. The bounding boxes are used for ROI classification and region-aware report generation training. Each of these segmentation masks and bounding boxes is paired with a specific text description, providing the biomedical classes or diagnostic findings of the annotated biomedical objects. 
	To ensure high-quality text descriptions, we extract semantic labels, diagnostic findings, medical knowledge, and clinical reports from publicly available datasets following previous methods~\cite{biomedparse,medtrinity,ctrate}. These descriptions capture fine-grained biomedical information across multiple granularities, including organs, lesions, tissues, and cells, enabling comprehensive interpretation of biomedical targets.
	
	A critical step in our dataset pre-processing pipeline is transforming the textual descriptions into a uniform Vision-Question-Answering (VQA) format~\cite{llava,Llava-med,sa2va,lisa}, aiming to facilitate universal training. For example, given a CT image with liver tumors, we define the question as: \emph{``Can you identify any abnormality within this CT image? Please respond with segmentation masks.''}. Then the corresponding answer is structured as: \emph{``It is [SEG]. Liver tumor''}. For the ROI classification task without segmentation outputs, the answer will only contain the biomedical class without the \emph{``[SEG]''} token~\cite{lisa,sa2va}.
	This uniform VQA format enables UniBiomed to jointly recognize abnormalities and segment biomedical targets within a universal foundation model. Some VQA examples are shown in \textbf{Extended Data Figure~\ref{fig_extended_vqa_format}}.
	
	Through this process, we construct a large-scale dataset consisting of \textbf{27 million} image-text-annotation triplets across 10 modalities, as shown in \textbf{Figure~\ref{fig_overview}~(a) and (b)}. Specifically, the 3D medical images (CT and MRI) are pre-processed as 2D slices following previous methods~\cite{medsam,biomedparse}. 
	To the best of our knowledge, this is the largest and most comprehensive dataset for biomedical grounded interpretation. For example, compared with the dataset used in the representative biomedical foundation model BiomedParse~\cite{biomedparse}, our dataset is over 30 times larger in scale. Unleashing the power of this large-scale dataset, UniBiomed demonstrates state-of-the-art performance across diverse tasks, outperforming existing biomedical foundation models~\cite{medsam,biomedparse,segvol,SAT,Llava-med,MedRegA,MedPLIB} by a large margin. The details of the used datasets are presented in \textbf{Extended Data Tables~\ref{table:dataset}, \ref{table:dataset_language}, \ref{table_dataset_descriptions}, and \ref{table_dataset_descriptions_modality}}. 
	
	\textbf{Method.} Multi-modal Large Language Models (MLLMs) have achieved remarkable effectiveness in processing vision-language information~\cite{llava,clip,internvl_original,Qwen-VL,blip}. Typically, MLLMs leverage a vision encoder to encode images as vision tokens and input them into a Large Language Model (LLMs)~\cite{llama} for text generation~\cite{llava}. However, state-of-the-art MLLMs~\cite{llava,clip,internvl_original,Qwen-VL} still fail to output the localizations of the corresponding regions described in the generated texts. This task, known as grounded interpretation in computer vision~\cite{liu2024grounding,li2022grounded,zou2023segment,lisa,sa2va,zhang2024omg}, requires a strong segmentation model to segment the target regions. Segment Anything Model (SAM)~\cite{sam,sam2} has demonstrated promising results in segmenting diverse sources of images, including biomedical images~\cite{medsam,biomedparse}. The advances in MLLMs~\cite{llava,clip,internvl_original,Qwen-VL} and SAM~\cite{sam,sam2} lead to a promising way for us to investigate grounded biomedical image interpretation.
	
	In this work, we introduce UniBiomed, which innovatively integrates MLLM~\cite{llava,internvl} and SAM~\cite{sam,sam2} for universal grounded interpretation of biomedical images. The overall framework is shown in \textbf{Figure~\ref{fig_overview}~(c)}. Our approach combines the complementary strengths of MLLM and SAM: MLLM processes both visual and textual information to generate descriptive interpretations, while SAM performs precise image segmentation through its vision encoder, prompt encoder, and mask decoder modules.
	
	The framework operates as an end-to-end way that, given user instructions and biomedical images, UniBiomed simultaneously: (1) generates text descriptions of the biomedical images, and (2) segments the corresponding biomedical targets.  
	The effectiveness of UniBiomed arises from the complementary integration of MLLM and SAM. 
	Specifically, UniBiomed combines the output of the MLLM~\cite{lisa,sa2va,glamm} with the tokenized user instructions as the language embeddings to guide SAM's segmentation process. 
	This incorporation effectively eliminates the manual efforts of crafting bounding boxes~\cite{medsam,segvol} or detailed text descriptions~\cite{SAT,m3d} as prompts for segmentation. 
	For implementation, we select InternVL2.5~\cite{internvl,internvl_original} as our foundation MLLM due to its robust multi-modal understanding capabilities, and SAM2~\cite{sam2} is adopted as our segmentation model for its improved segmentation performance. The details of the method are described in \textbf{Section~\ref{sec_method_framework}}.
	
	\textbf{Evaluation}. To demonstrate the effectiveness of UniBiomed on diverse biomedical tasks, we conduct a comprehensive evaluation on 70 internal and 14 external datasets. Since we focus on evaluating the effectiveness of grounded interpretation, for comparison methods, we compare segmentation foundation models~\cite{medsam,biomedparse,segvol,SAT} on segmentation tasks, compare grounding tasks with grounding methods~\cite{lisa,glamm}, and compare medical MLLMs~\cite{Llava-med,MedRegA,MedPLIB} on medical diagnosis tasks. \textbf{(1)} First, we conduct extensive comparisons with several foundation models~\cite{medsam,biomedparse,segvol,SAT} in biomedical image segmentation. \textbf{(2)} Second, we evaluate UniBiomed in a challenging grounded VQA task, \emph{i.e.}, grounded disease recognition, which aims to simultaneously generate diagnostic findings and segment the corresponding biomedical targets. \textbf{(3)} Then, we conduct extensive experiments on grounded report generation, which is designed to generate clinical reports of the given images and highlight the target regions. We verify the grounded report generation task on the RadGenome~\cite{radgenome,ctrate} dataset. \textbf{(4)} Finally, we verify the effectiveness of UniBiomed in two region-aware diagnosis tasks: ROI classification and region-aware report generation. 
	Specifically, we extract bounding boxes from segmentation masks to indicate the ROIs following previous methods~\cite{MedPLIB,osprey,vipllava,ROI_report}. We assess the region-aware report generation task using the MedTrinity~\cite{medtrinity} dataset. The details of the datasets are presented in \textbf{Extended Data Tables~\ref{table:dataset}, \ref{table:dataset_language}, \ref{table_dataset_descriptions}, and \ref{table_dataset_descriptions_modality}}. More details of the evaluation are presented in \textbf{Section~\ref{sec_method_evaluation}}.
	
	\subsection{UniBiomed excels in biomedical image segmentation}
	
	In this section, we present a comprehensive comparison with multiple biomedical segmentation foundation models in biomedical image segmentation, \emph{i.e.}, MedSAM~\cite{medsam}, SegVol~\cite{segvol}, SAT~\cite{SAT}, and BiomedParse~\cite{biomedparse}, as shown in \textbf{Figure~\ref{fig_segmentation}}. Specifically, we evaluate our method on 46 internal and 14 external datasets across nine diverse biomedical imaging modalities. 
	The details of the datasets are shown in \textbf{Extended Data Tables~\ref{table:dataset}}. 
	We adopt text prompts for referring segmentation as SAT~\cite{SAT} and BiomedParse~\cite{biomedparse}. For example, to segment liver tumors in CT, we provide a text instruction \emph{``Please segment liver tumor in the CT image''} to the model, eliminating the need to offer visual prompts like bounding boxes or points as MedSAM~\cite{medsam} and SegVol~\cite{segvol}. More details of the competing methods are presented in \textbf{Section~\ref{sec_method}}.
	
	As shown in \textbf{Figure~\ref{fig_segmentation} (a) and (b)}, UniBiomed achieves superior performance compared to state-of-the-art segmentation foundation models~\cite{medsam,segvol,SAT,biomedparse}. Specifically, UniBiomed surpasses the best-competing method BiomedParse~\cite{biomedparse} by a clear margin, \emph{i.e.}, $10.25\%$ dice score improvements on 60 datasets across nine modalities. \textbf{Figure~\ref{fig_segmentation} (d)} shows that UniBiomed outperforms BiomedParse by $9.13\%$ and $13.95\%$ in 46 internal and 14 external datasets, respectively. 
	These findings robustly underscore UniBiomed’s substantial advancement in universal biomedical image segmentation.
	
	One of the key reasons for UniBiomed’s breakthrough in biomedical image segmentation lies in its ability to unleash the power of MLLM and SAM in large-scale multi-modal biomedical datasets. Unlike prior segmentation models~\cite{medsam,segvol,SAT,biomedparse} that rely solely on segmentation datasets for training, UniBiomed effectively incorporates large-scale VQA and report generation datasets~\cite{ctrate,medtrinity} in universal training. This approach enables UniBiomed to develop more generalizable representations, leading to superior segmentation performance across diverse biomedical imaging modalities. As shown in \textbf{Figure~\ref{fig_segmentation}~(e)}, with large-scale multi-modal datasets for training, we improve UniBiomed by $3.39\%$ and $8.26\%$ dice scores in internal and external validation, respectively. 
	
	More importantly, beyond segmentation, UniBiomed establishes a more comprehensive interpretation capability for biomedical image analysis, unlocking novel opportunities in broader clinical applications. As illustrated in \textbf{Figure~\ref{fig_segmentation} (f)}, prior segmentation models~\cite{medsam,segvol,SAT,biomedparse} are still limited in tackling the segmentation tasks, which heavily hamper their clinical applications in biomedical image analysis. In contrast, UniBiomed extends segmentation to grounded interpretation with text generation, which is a versatile foundation model capable of handling diverse biomedical tasks, making it more adaptable for clinical applications.

	\begin{figure*}
		\centering
		\includegraphics[width=1\linewidth]{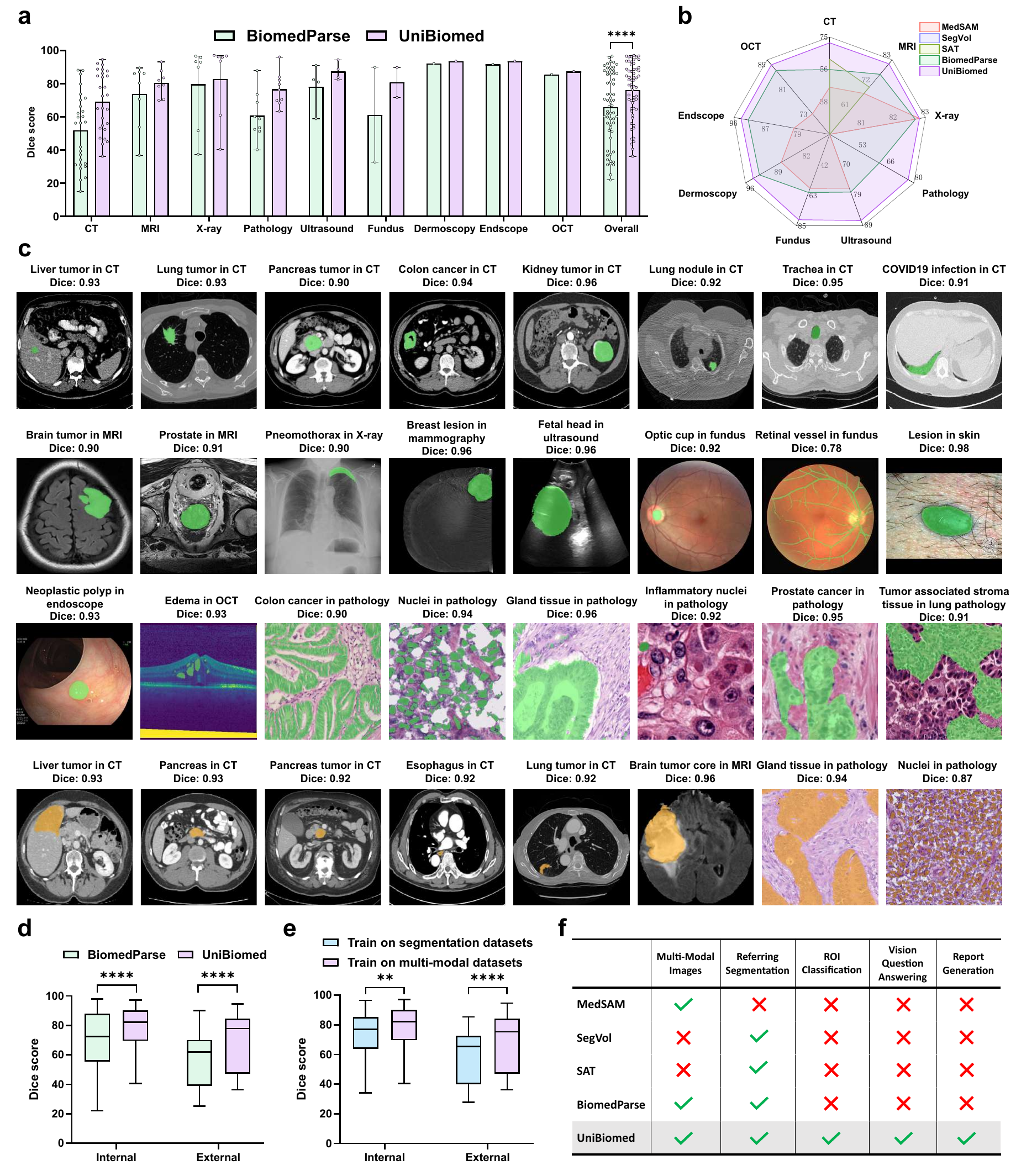}
	\end{figure*}
	\clearpage
	\begin{figure*}
		\centering
		\caption{
			\textbf{Comparison of biomedical image segmentation}. \textbf{a.} Comparison of Dice scores between UniBiomed and the best-competing segmentation foundation model BiomedParse~\cite{biomedparse} across nine modalities, with the mean ($\pm$s.d.) reported. Significance levels at which UniBiomed outperforms BiomedParse~\cite{biomedparse}, with a two-sided paired t-test $P$-value of ****$P<1\times10^{-4}$. Notably, UniBiomed outperforms BiomedParse~\cite{biomedparse} by an average of $10.25\%$ in dice scores on 60 internal and external datasets. \textbf{b.} Radar chart comparisons with several representative segmentation foundation models, MedSAM~\cite{medsam}, SegVol~\cite{segvol}, SAT~\cite{SAT}, and BiomedParse~\cite{biomedparse}. Specifically, SAT~\cite{SAT} is only applicable to 3D modalities like CT and MRI. SegVol~\cite{segvol} is solely available for CT datasets. \textbf{c.} Qualitative segmentation results of UniBiomed. Specifically, the colors of segmentation masks for internal and external validations are \textcolor{green}{green} and \textcolor{orange}{orange}, respectively. \textbf{d.} Box plot comparisons between UniBiomed and BiomedParse~\cite{biomedparse}. Significance levels at which UniBiomed outperforms BiomedParse~\cite{biomedparse}, with two-sided paired t-test $P$-values of ****$P<1\times10^{-4}$ for both internal and external validations. 
			\textbf{e.} Ablation studies of UniBiomed. We compare UniBiomed under two settings: (1) train on only segmentation datasets, and (2) train on large-scale multi-modal datasets, including segmentation, VQA, and report generation datasets. It can be seen that with large-scale multi-modal datasets for training, the segmentation performance can be further improved, the two-sided paired t-test $P$-values are **$P<1\times10^{-2}$ and ****$P<1\times10^{-4}$ for internal and external validations, respectively.
			\textbf{f.} Overall application comparisons with representative biomedical segmentation models~\cite{medsam,segvol,SAT,biomedparse}. Referring segmentation represents using text instructions as prompts for segmentation, which eliminates the efforts of providing tight bounding boxes for the biomedical objects~\cite{medsam}. Notably, UniBiomed not only supports multi-modal biomedical image segmentation but also establishes remarkable capabilities in ROI classification, VQA, and report generation, where state-of-the-art segmentation models~\cite{medsam,segvol,SAT,biomedparse} fail to establish.
		}
		\label{fig_segmentation}
	\end{figure*}

	\subsection{UniBiomed enables accurate grounded disease recognition}
	
	In medical MLLMs~\cite{Biomedgpt,RADFM,Llava-med,ctrate,m3d,fvlm,gsco}, disease recognition is one of the most important VQA tasks, which aims to identify the lesions, tumors, or abnormalities in the given images. However, existing medical MLLMs~\cite{Biomedgpt,RADFM,Llava-med,ctrate,m3d,fvlm,gsco} are typically region-agnostic, which solely generate diagnostic findings but fail to localize relevant regions (\emph{e.g.}, tiny lesions) in the images. This lack of spatial awareness severely limits their clinical applications, as precise localization of abnormalities is essential for diagnosis and treatment planning.
	
	In this work, we introduce a novel and challenging grounded VQA task, \emph{i.e.}, grounded disease recognition, which is designed to simultaneously generate diagnostic findings and segment the corresponding targets. In this task, the model is asked to predict abnormality class (\emph{e.g.}, liver tumor or pancreas tumor) and segment the targets in the images. If there is no abnormality in the given images, the model is supposed to output \emph{``No findings''} and produce all-zero segmentation masks. 
	
	\begin{figure*}[!h]
		\centering
		\includegraphics[width=1\linewidth]{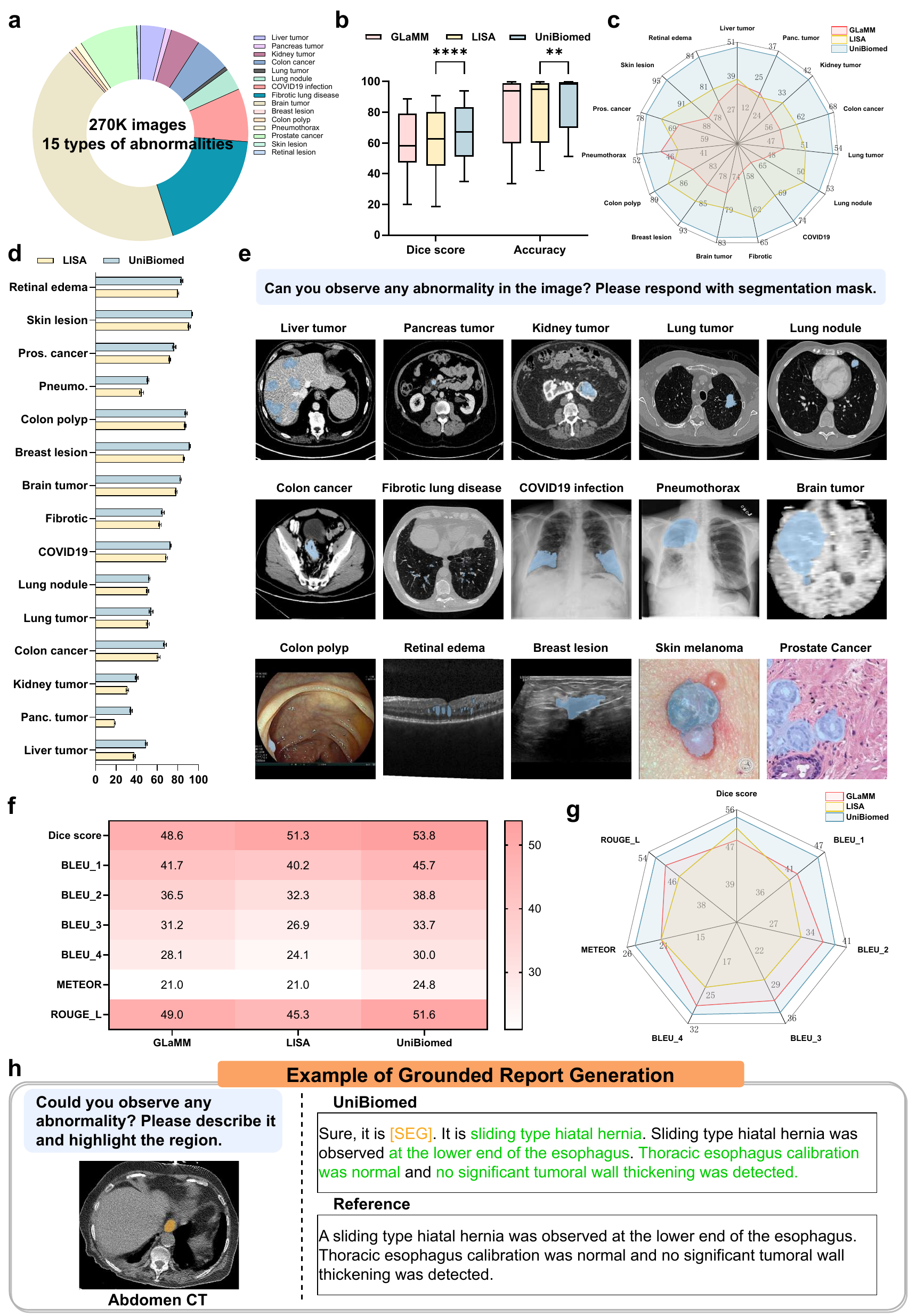}
	\end{figure*}
	\clearpage
	\begin{figure*}
		\centering
		\caption{
			\textbf{Comparison of grounded disease recognition and grounded report generation}. 
			\textbf{a.} We curate a new dataset for grounded disease recognition. Specifically, this dataset contains 270K images with 15 types of abnormalities. The details of the dataset are presented in \textbf{Extended Data Table~\ref{table:dataset_grounded_disease}}.
			\textbf{b.} We compare UniBiomed with two state-of-the-art methods, LISA~\cite{lisa} and GLaMM~\cite{glamm}, in segmentation dice scores and disease recognition accuracy. Notably, UniBiomed outperforms the best competing method LISA~\cite{lisa} by $3.86\%$ in dice scores and $3.29\%$ in accuracy, with two-sided paired t-test $P$-values of ****$P<1\times10^{-4}$ and **$P<1\times10^{-2}$. 
			\textbf{c.} Radar chart comparisons across 15 types of abnormalities. \textbf{d.} Dice scores comparisons between UniBiomed and LISA~\cite{lisa} across 15 types of abnormalities, with the mean ($\pm$s.d.) reported. \textbf{e.} The qualitative visualization results of grounded disease recognition. The template user instruction is shown in the blue box. Given template instructions, UniBiomed can recognize the abnormalities in the images and highlight the precise localizations. \textbf{f.} Comparisons of grounded report generation on the RadGenome~\cite{radgenome} dataset. We re-implement GLaMM~\cite{glamm} and LISA~\cite{lisa} on this dataset for comparison. We report dice scores results for segmentation evaluation, BLEU~\cite{bleu}, METEOR~\cite{meteor}, and ROUGE\_L~\cite{rouge} results for report generation evaluation. \textbf{g.} Radar chart comparisons across different metrics in grounded report generation. \textbf{h.} An example of grounded report generation on abdominal CT images. The template user instruction is shown in the blue box. The text in \textcolor{mygreen}{green} indicates the correct contents. Reference denotes the ground truth from the RadGenome dataset~\cite{radgenome}. The segmentation mask indicates the location of the corresponding organ (esophagus in the example) described in the report. 
		}
		\label{fig_grounded}
	\end{figure*}
	
	To this end, we curate a comprehensive dataset encompassing 15 distinct abnormality types for model training and validation, as shown in \textbf{Figure~\ref{fig_grounded} (a)}. Specifically, these 15 abnormality types contain liver tumor, pancreas tumor, kidney tumor, colon cancer, lung tumor, lung nodule, COVID-19 infection, fibrotic lung disease, brain tumor, breast lesion, colon polyp, pneumothorax, prostate cancer, skin lesion, and retinal lesion. The details of the datasets are shown in \textbf{Extended Data Tables~\ref{table:dataset} and \ref{table:dataset_grounded_disease}}.
	
	Existing medical models lack the capability to tackle this task, which requires simultaneously outputting segmentation masks and diagnostic findings. Although BiomedParse~\cite{biomedparse} can conduct segmentation and employ a meta-object classifier for classification, this classifier requires users to pre-diagnose the images and provide diagnostic findings as textual prompts to the model, thus failing to perform the grounded disease recognition task introduced in this work. To this end, we re-implement two state-of-the-art approaches from the general domain, LISA~\cite{lisa} and GLaMM~\cite{glamm}, on our curated datasets for fair comparisons. The comparison results are shown in \textbf{Figure~\ref{fig_grounded} (b-d)}. Specifically, UniBiomed outperforms the best competing method LISA~\cite{lisa} by $3.86\%$ and $3.29\%$ in segmentation dice scores and disease recognition accuracy, respectively. Consistent improvements across different types of abnormalities are observed in \textbf{Figure~\ref{fig_grounded}~(d)}, which robustly validates the effectiveness of UniBiomed in this challenging VQA task.
	
	
	More importantly, the strong capability of UniBiomed in grounded disease recognition potentially leads to a significant paradigm shift in the biomedical image analysis workflow. Specifically, previous biomedical foundation models heavily rely on clinical experts to pre-diagnose images and manually craft precise textual or visual prompts. 
	For example, for the task of lung tumor recognition, one of the representative foundation models, MedSAM~\cite{medsam}, requires radiologists to outline the regions of lung tumors in each slice of a 3D CT scan and provide the prompts to the model. Another representative foundation model, BiomedParse~\cite{biomedparse}, necessitates radiologists to identify lung tumors in each slice of a 3D CT scan in advance, then input an instruction \emph{``Please segment lung tumor in this CT image''} to prompt the model for lung tumor segmentation. In contrast, UniBiomed enables an end-to-end pipeline to recognize abnormalities in images. As shown in \textbf{Figure~\ref{fig_grounded}~(e)}, UniBiomed is free of pre-diagnosis by clinical experts. Instead, UniBiomed adopts a template instruction and automatically recognizes the abnormality class with segmentation predictions. These findings underscore the clinical practice of UniBiomed's grounded interpretation capabilities, which is a non-trivial advancement compared to state-of-the-art biomedical foundation models~\cite{medsam,segvol,SAT,biomedparse}.

	\subsection{UniBiomed enables accurate grounded report generation}
	
	Grounded report generation is a challenging task that combines biomedical image segmentation with report generation, enabling end-to-end biomedical image analysis in a universal model. Unlike prior methods~\cite{biomedparse,MedRegA,Llava-med,MedPLIB}, which rely on separate segmentation models and standalone language models for report generation, UniBiomed performs both tasks within one universal model. This integration allows UniBiomed to leverage holistic biomedical knowledge more effectively, improving both segmentation and report quality.
	
	To evaluate its effectiveness, we conduct extensive comparisons on the RadGenome~\cite{radgenome} dataset. Specifically, the RadGenome dataset is processed from the CT-RATE~\cite{ctrate} dataset, which contains detailed CT reports with segmentation masks for training and validation. The clinical reports provide detailed descriptions of the images, and the segmentation masks indicate the localization of corresponding organs. Since existing medical MLLMs~\cite{Biomedgpt,Llava-med,RADFM,gsco,m3d,fvlm,MedRegA,MedPLIB} cannot tackle this task, we also re-implement GLaMM~\cite{glamm} and LISA~\cite{lisa} on our datasets for comparisons. The results are shown in \textbf{Figure~\ref{fig_grounded} (f) and (g)}. 
	
	We report the dice scores results for segmentation evaluation, Bilingual Evaluation Understudy (BLEU)~\cite{bleu}, Metric for Evaluation of Translation with Explicit ORdering (METEOR)~\cite{meteor}, and Recall-Oriented Understudy for Gisting Evaluation (ROUGE\_L)~\cite{rouge} results for report generation evaluation. The details of the evaluation metrics are described in \textbf{Section~\ref{sec_method_evaluation}}. It can be observed that UniBiomed not only achieves better segmentation performance ($53.8\%$ dice scores) but also demonstrates superior report generation capabilities ($45.7\%$, $24.8\%$, and $51.6\%$ in BLEU\_1, METEOR, and ROUGE\_L, respectively). We further showcase an example of grounded report generation on abdomen CT images, as presented in \textbf{Figure~\ref{fig_grounded} (h)}. Specifically, within an end-to-end process, UniBiomed can observe the \emph{``sliding type hiatal hernia''} abnormality with detailed descriptions and highlight the anatomy region of the esophagus with segmentation masks. In contrast, state-of-the-art methods~\cite{medsam,biomedparse,segvol,SAT,Biomedgpt,Llava-med,MedRegA,MedPLIB} require multiple models and stages to establish this diagnosis process, which is inflexible and inefficient in real-world deployment. More examples are presented in \textbf{Extended Data Figure~\ref{fig_extended_grg}}.
	
	\subsection{UniBiomed effectively improves region-aware diagnosis}
	
	Region-aware diagnosis refers to recognizing the biomedical objects within the user-defined ROIs in the images~\cite{medtrinity,MedPLIB}. We involve this task in our training to enhance the region-aware ability of UniBiomed and ultimately improve the performance of grounded interpretation. Following the settings of MedPLIB~\cite{MedPLIB}, we evaluate two region-aware diagnosis tasks in this work, \emph{i.e.}, ROI classification and region-aware report generation. Since the segmentation predictions are not required in this task, we output the results by the MLLM module only. 
	Following previous methods~\cite{medtrinity,vipllava}, to indicate the ROIs, we directly overlay bounding boxes onto the images, eliminating the need for complex region encodings for indicating ROIs. The details are presented in \textbf{Section~\ref{sec_method}}.
	
	We first evaluate the performance of ROI classification. 
	For fair comparisons, we compare UniBiomed with two state-of-the-art medical MLLMs, MedRegA~\cite{MedRegA} and MedPLIB~\cite{MedPLIB}, since both of them were trained on large-scale ROI classification datasets. 
	The results are shown in \textbf{Figure~\ref{fig_roi}~(a-d)}. Specifically, UniBiomed achieves an average $93.38\%$ accuracy in ROI classification, surpassing the best competing method MedPLIB~\cite{MedPLIB} by an average of $8.32\%$ in ROI classification accuracy across ten imaging modalities. These results validate that UniBiomed also excels in ROI classification compared with state-of-the-art methods.
	
	\begin{figure*}[!h]
		\centering
		\includegraphics[width=0.97\linewidth]{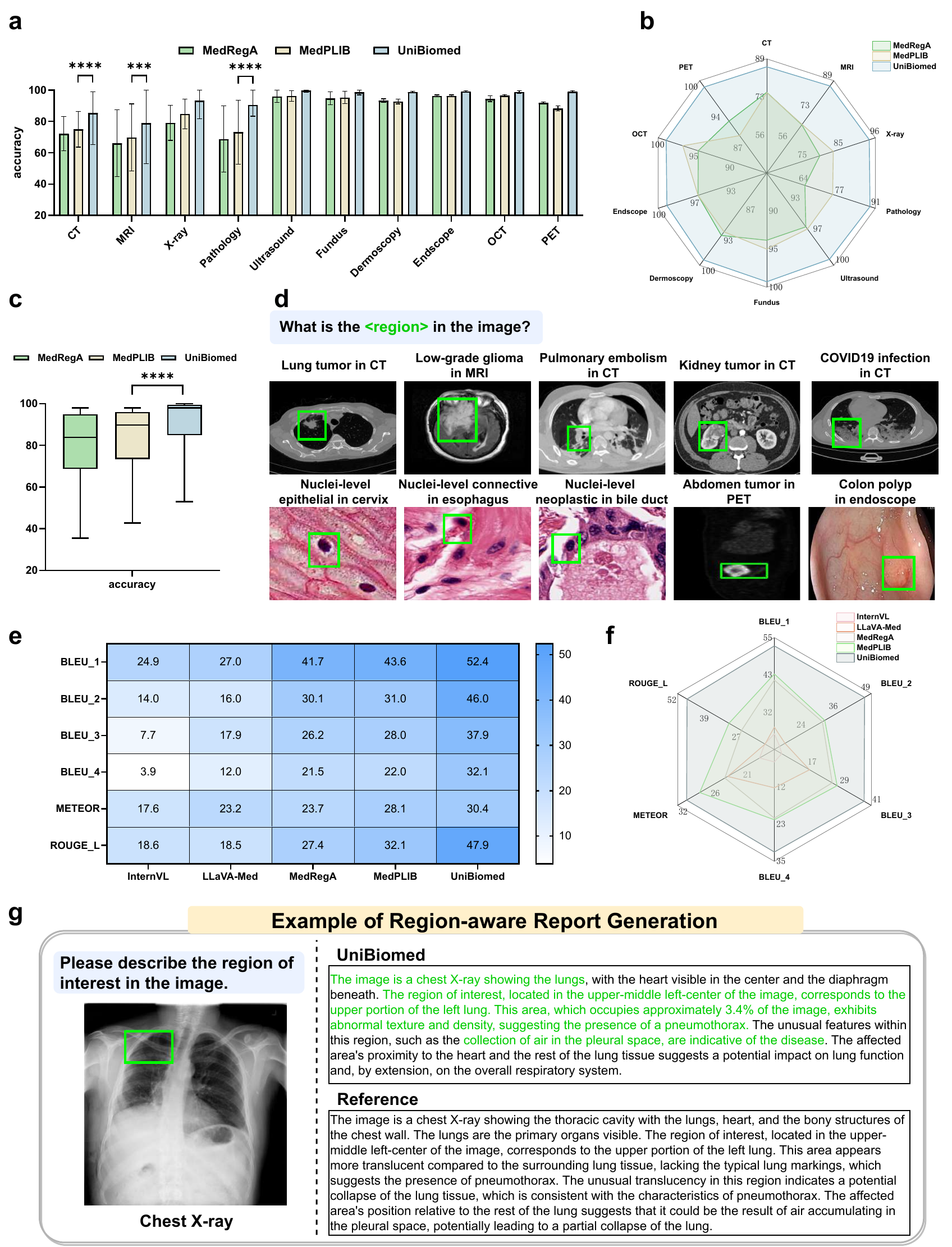}
	\end{figure*}
	\clearpage
	\begin{figure*}
		\centering
		\caption{
			\textbf{Comparison of ROI classification and region-aware report generation}.
			\textbf{a.} Comparisons with MedRegA~\cite{MedRegA} and MedPLIB~\cite{MedPLIB} in ROI classification, with the mean accuracy ($\pm$s.d.) reported. Significance levels at which UniBiomed outperforms MedPLIB~\cite{MedPLIB}, with two-sided paired t-test $P$-values are ****$P<1\times10^{-4}$, ***$P<1\times10^{-3}$, and ****$P<1\times10^{-4}$ for CT, MRI, and pathology, respectively. 
			\textbf{b.} Radar chart comparison with MedRegA~\cite{MedRegA} and MedPLIB~\cite{MedPLIB} across ten diverse biomedical imaging modalities. \textbf{c.} UniBiomed surpasses MedPLIB~\cite{MedPLIB} by $8.32\%$ in ROI classification accuracy, with two-sided paired t-test $P$-values of ****$P<1\times10^{-4}$. \textbf{d.} Qualitative visualization results of ROI classification. The template user instruction is shown in the blue box. Specifically, given a bounding box prompt, UniBiomed can effectively predict the class of biomedical targets within ROIs.
			\textbf{e.} Comparisons on region-aware report generation in the MedTrinity~\cite{medtrinity} dataset. We compare UniBiomed with multiple representative MLLMs, including InternVL2.5~\cite{internvl,internvl_original}, LLaVA-Med~\cite{Llava-med}, MedRegA~\cite{MedRegA}, and MedPLIB~\cite{MedPLIB}. We report the results of BLEU~\cite{bleu}, METEOR~\cite{meteor}, and ROUGE\_L~\cite{rouge}. \textbf{f.} Radar chart comparisons across different metrics in region-aware report generation. \textbf{g.} An example of region-aware report generation on chest x-ray image. The template user instruction is shown in the blue box. The text in \textcolor{mygreen}{green} indicates the correct contents. Reference denotes the ground truth from the MedTrinity dataset~\cite{medtrinity}.}
		\label{fig_roi}
	\end{figure*}
	
	We further evaluate a more challenging region-aware diagnosis task, \emph{i.e.}, region-aware report generation, which aims to generate detailed reports within ROIs in the biomedical images. Our experiments are conducted on the MedTrinity~\cite{medtrinity} dataset, a large-scale benchmark featuring region-centric reports across diverse biomedical imaging modalities. Specifically, MedTrinity is aggregated from 23 public VQA and report generation datasets across 10 biomedical imaging modalities, as described in \textbf{Extended Data Table~\ref{table:dataset_language}}.
	We compare our approach with several state-of-the-art MLLMs, including InternVL2.5~\cite{internvl,internvl_original}, LLaVA-Med~\cite{Llava-med}, MedRegA~\cite{MedRegA}, and MedPLIB~\cite{MedPLIB}. Among them, InternVL2.5 is the state-of-the-art MLLM in the general domain. LLaVA-Med, MedRegA, and MedPLIB are generalist MLLMs in the medical domain, while MedRegA and MedPLIB were trained on large-scale region-centric report generation datasets. We report the results of BLEU~\cite{bleu}, METEOR~\cite{meteor}, and ROUGE\_L~\cite{rouge}. The details of the comparison methods are described in \textbf{Section~\ref{sec_method_evaluation}}. 
	
	The results are shown in \textbf{Figure~\ref{fig_roi} (e) and (f)}. It can be seen that UniBiomed achieves superior performance compared with previous MLLMs~\cite{internvl,Llava-med,MedRegA,MedPLIB}. Specifically, UniBiomed achieves $52.4\%$, $30.4\%$, and $47.9\%$ in BLEU\_1, METEOR, and ROUGE\_L, respectively, surpassing the best-competing method MedPLIB~\cite{MedPLIB} by $8.8\%$, $2.3\%$, and $15.8\%$. These results robustly validate the effectiveness of UniBiomed in this task.
	We further present an example of region-aware report generation on prostate pathology images, as shown in \textbf{Figure~\ref{fig_roi} (g)}. It can be seen that UniBiomed can generate a detailed description to illustrate the observed features.  
	Specifically, UniBiomed not only accurately identifies the abnormality (\emph{i.e.}, pneumothorax) within the target regions, but also generates a detailed description to illustrate the observed features. 
	More examples are presented in \textbf{Extended Data Figure~\ref{fig_extended_roi_report}}.
	
	\subsection{UniBiomed optimizes biomedical image analysis workflow}
	
	\begin{figure*}
		\centering
		\includegraphics[width=1\linewidth]{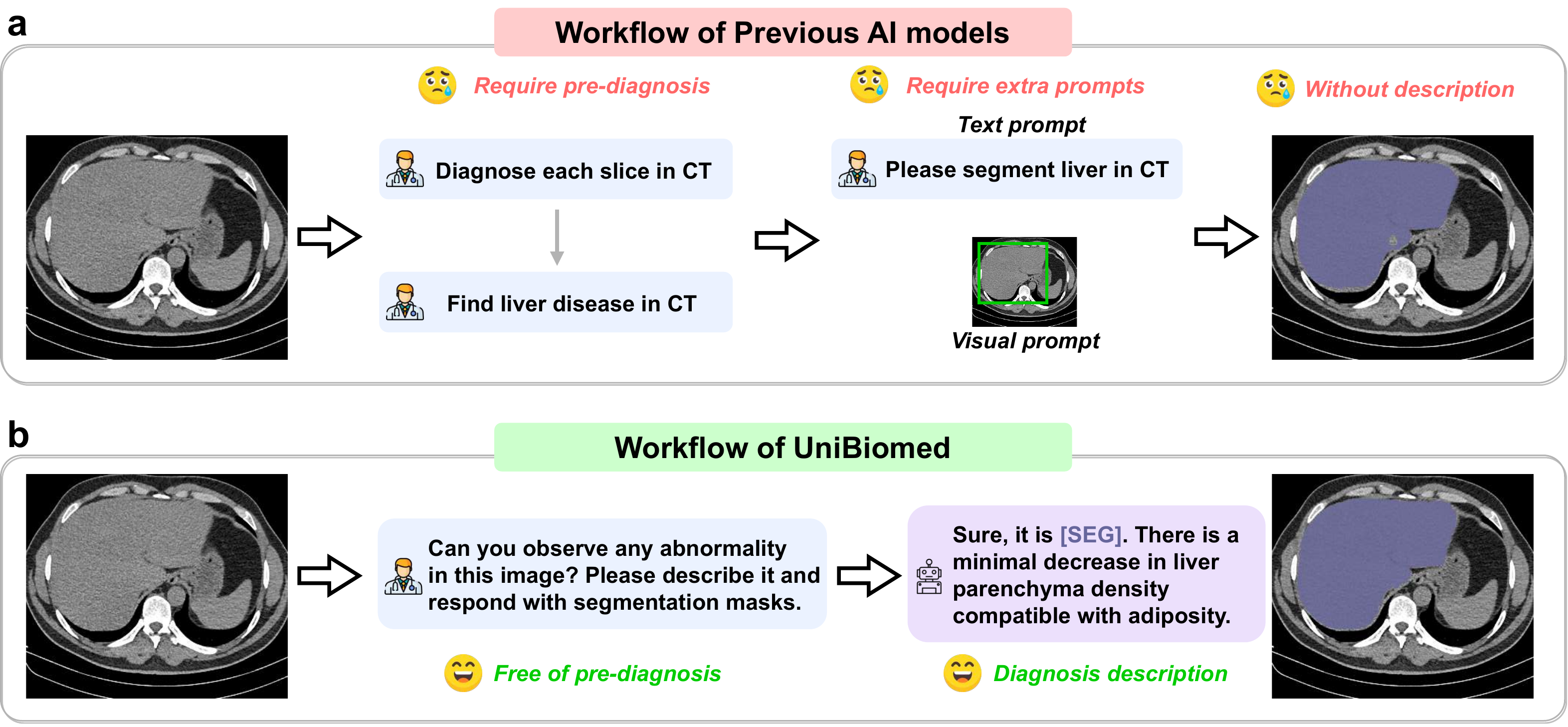}
		\caption{\rebut{\textbf{Workflow comparisons}. a presents the workflow of previous biomedical segmentation foundation models, \emph{e.g.}, MedSAM~\cite{medsam} and BiomedParse~\cite{biomedparse}. b shows the workflow of our introduced UniBiomed. We present an example of grounded report generation on CT images.}}
		\label{fig_workflow}
	\end{figure*}
	
	In this section, we show that UniBiomed is an effective AI tool for improving the efficiency of the biomedical image analysis workflow. 
	As shown in \rebut{\textbf{Figure~\ref{fig_workflow}a}}, previous biomedical segmentation foundation models such as MedSAM~\cite{medsam} and BiomedParse~\cite{biomedparse} follow a cumbersome process. Given an input CT scan, these models require radiologists to first pre-diagnose each slice to locate target abnormalities, such as liver disease in the example. Once identified, additional manual inputs are necessary, \emph{e.g.}, BiomedParse~\cite{biomedparse} demands a text description as the textual prompt, while MedSAM~\cite{medsam} relies on a tightly drawn bounding box prompt. More critically, these models are limited to segmentation tasks and fail to generate detailed diagnostic descriptions of the targets. These inefficiencies significantly hinder their clinical applications. 
	
	In contrast, UniBiomed overcomes these limitations through a novel integration of MLLM and SAM. Given an input CT scan, UniBiomed automatically detects abnormalities, generates precise segmentation masks, and provides detailed diagnostic descriptions. There are two significant advantages in this workflow: (1) First, this process is free of pre-diagnosis by radiologists and requires no manual prompts, eliminating the time-consuming process of slice-by-slice analysis and manual prompt engineering. (2) Moreover, unlike prior foundation models~\cite{medsam,biomedparse,segvol,SAT}, which are limited to segmentation outputs, UniBiomed is capable of combining visual analysis with diagnostic text generation. For example, UniBiomed not only segments targets but also delivers clinically relevant interpretations, such as identifying \emph{``a minimal decrease in liver parenchyma density compatible with adiposity''} as shown in the example.
	These advancements lead to a significant paradigm shift in biomedical image analysis, which optimizes workflow efficiency and enables an end-to-end diagnosis workflow.

	\section{Discussion}
	
	In this work, we introduce UniBiomed, the first universal foundation model for grounded biomedical image interpretation. UniBiomed effectively unifies the generation of diagnostic findings with the segmentation of corresponding biomedical targets, enabling accurate biomedical image analysis across ten diverse imaging modalities. UniBiomed is based on a novel integration of advanced Multi-modal Large Language Model (MLLM) and Segment Anything Model (SAM)~\cite{sam,sam2}, which can effectively leverage multi-modal biomedical information for tackling diverse biomedical tasks. To develop UniBiomed, we curate a large-scale dataset comprising 27 million triplets of images, region annotations, and text descriptions spanning ten biomedical imaging modalities, which is the largest and most comprehensive dataset for biomedical grounded interpretation. To evaluate the effectiveness of UniBiomed, we conduct a large-scale validation on 84 datasets with comprehensive comparisons of state-of-the-art methods~\cite{medsam,biomedparse,segvol,SAT,Llava-med,lisa,glamm,internvl,MedRegA,MedPLIB}. \rebut{To evaluate the generalizability of our model, we involve 14 external datasets in validation, which are unseen in the training.} Extensive experiments demonstrate that UniBiomed achieves state-of-the-art performance on a wide range of biomedical tasks, including biomedical image segmentation, disease recognition, region-aware diagnosis, and report generation. We further showcase that UniBiomed represents a novel paradigm shift in biomedical image analysis workflows, which effectively improves the efficiency of AI-assisted biomedical image analysis.
	These findings demonstrate the promising prospects of UniBiomed in clinical applications. 
	
	Recent advances have attracted increasing attention to multimodal information for biomedical image analysis~\cite{Biomedgpt,Llava-med,gsco,RADFM,fvlm,m3d,nie2025conceptclip,nie2025explainable}. Concretely, visual features extracted from biomedical images provide anatomical and functional information from cell to organ levels~\cite{nnUNet,biomedparse,xu2024whole,sun2024foundation,wu2024freetumor,zhuang2024mim,wu2025freetumor,msd}, while textual descriptions from clinical experts offer fine-grained information for interpreting biomedical images~\cite{Biomedgpt,peng2023ai,ni2024mg,lu2024multimodal}. Although recent biomedical AI models have showcased encouraging results in biomedical image analysis, it remains challenging to effectively integrate holistic vision and language information for assisting clinicians in practice. Specifically, the clinical adoption of biomedical AI models requires that the AI-generated results are both accurate and interpretable for clinicians to ensure reliability. Thus, it demands that biomedical AI models can output the diagnostic findings and simultaneously highlight the corresponding regions in the images, offering rich visual and textual information for assisting clinicians.
	
	To this end, we underscore the importance of grounded interpretation in biomedical image analysis, \emph{i.e.}, generating diagnostic findings and simultaneously segmenting the corresponding biomedical objects. This task enables models to extract important biomedical targets (\emph{e.g.}, lesions, tumors, cancer cells, or abnormal organs) and generate detailed descriptions (\emph{e.g.}, clinical reports) of corresponding objects to assist clinical experts in diagnosis, unlocking novel opportunities for end-to-end biomedical image analysis.
	
	However, existing biomedical models~\cite{nnUNet,medsam,biomedparse,segvol,SAT,medsam2,Llava-med,lisa,glamm,fvlm,m3d,RADFM,MedRegA,MedPLIB} fail to tackle this challenging yet critical task. In particular, grounded interpretation requires simultaneous execution of segmentation and text generation, whereas current models can only conduct these two tasks independently. Concretely, the biomedical segmentation foundation models~\cite{medsam,biomedparse,segvol,SAT,medsam2} can deliver segmentation masks while failing to establish text generation tasks such as VQA and report generation. Although recent medical MLLMs~\cite{Llava-med,fvlm,m3d,gsco,RADFM,MedRegA,MedPLIB} have demonstrated promising results in generating text descriptions for biomedical images, these models are not capable of segmenting the corresponding biomedical objects (\emph{e.g.}, tiny lesions) simultaneously, making it difficult for clinicians to identify the corresponding regions described in the reports. To illustrate the drawbacks of these methods more clearly, we present a detailed comparison in \textbf{Extended Data Table~\ref{table_novelty}}. 
	Compared with previous methods, UniBiomed is the first foundation model to support multi-modal image interpretation, flexible user prompts, segmentation prediction, text description generation, region-aware diagnosis, pixel-level grounding, and end-to-end training in one universal model. This breakthrough paves a promising path towards more accurate and efficient biomedical image analysis.
	
	The ability to leverage different types of biomedical datasets is pivotal to the success of UniBiomed. Previous biomedical foundation models are typically based on disjoint training on separate datasets, \emph{e.g.}, text generation datasets for language models~\cite{Biomedgpt,Llava-med} or segmentation datasets for segmentation models~\cite{medsam,biomedparse,stringer2021cellpose}. This limitation impedes their ability to leverage holistic vision and language information for biomedical image analysis. In contrast, within a universal training process, UniBiomed can simultaneously leverage image segmentation datasets, vision question answering datasets, report generation datasets, and region-aware diagnosis datasets for developing a strong biomedical foundation model. This distinguished advantage makes UniBiomed stand out from previous biomedical foundation models, achieving better performance across diverse biomedical tasks. We present ablation studies to evaluate the effectiveness of universal training in \textbf{Extended Data Figure~\ref{fig_extended_ablation}}.
	
	One of the major bottlenecks is the availability of data. To train UniBiomed, the training datasets necessitate both spatial localization and corresponding text descriptions for the biomedical objects. 
	Inspired by the recent advances of ``Visual Instruction Tuning'' in MLLMs~\cite{llava,lisa,sa2va}, we construct a uniform VQA format to facilitate universal grounded interpretation of biomedical images. In this way, we curate a large-scale dataset comprising 27 million image-text-annotation triplets across ten biomedical imaging modalities for training and validation, which is the largest and most comprehensive dataset in this field. Our curated dataset will fortify the foundation of future research in biomedical grounded interpretation.
	
	We further showcase that UniBiomed is a more practical biomedical AI tool for optimizing workflow efficiency. Prior segmentation foundation models~\cite{medsam,biomedparse,segvol,SAT} heavily rely on clinical experts to provide accurate textual or visual prompts. Take the representative models, MedSAM~\cite{medsam} and BiomedParse~\cite{biomedparse} as examples. Both of these models require the users to identify the targets of the input images in advance. Then, visual or textual prompts should be provided, \emph{e.g.}, MedSAM~\cite{medsam} requires tightly drawn bounding boxes as visual prompts while BiomedParse~\cite{biomedparse} necessitates text instructions from pre-diagnostic findings as textual prompts. This process is tedious for clinicians and significantly hampers workflow efficiency. For example, for 3D medical images like CT and MRI, there are probably a few hundred slices within a scan, while these models~\cite{medsam,biomedparse,segvol,SAT} require users to pre-diagnose the scans slice-by-slice and provide accurate prompts. In addition, these models~\cite{medsam,biomedparse,segvol,SAT} still fail to generate diagnostic findings for interpreting the images, which significantly hinders their clinical applications.
	In contrast, UniBiomed provides automated end-to-end grounded interpretation of biomedical images, with both precise segmentation predictions and detailed text descriptions, which significantly optimizes the workflow efficiency of biomedical image analysis. 
	
	Although promising results have been demonstrated by UniBiomed, there are still several limitations for further improvement. \rebut{Despite the extensive scale of our curated dataset, it may still lack incorporation of certain rare diseases or conditions, potentially limiting the model’s generalizability across all medical scenarios. This highlights the need for continuous dataset expansion and diversification to ensure comprehensive biomedical image analysis.}
	In addition, while UniBiomed has achieved satisfactory performance on large-scale public datasets, further exploration of its clinical application is necessary to substantiate the effectiveness of our method. Moving forward, we will work closely with clinical practitioners to ensure that the model addresses practical needs and integrates seamlessly into existing workflows.
	
	\rebut{
		\textbf{Discussion with medical agents.} Medical agents~\cite{medrax,li2024mmedagent,nath2025vila} have also received increasing attention in medical image analysis, which use LLM as the orchestrator and call different models to tackle various tasks. A key advantage of medical AI agents is their extensibility, which can be adapted for novel tasks by integrating new models. In contrast, universal foundation models~\cite{Biomedgpt,biomedparse,Llava-med}, including UniBiomed, can tackle various tasks by only one model, which is more flexible and also eliminates the need to interoperate between different models. Recent work~\cite{gsco} also demonstrated that a superior universal model can serve as a better orchestrator in medical agents. In the future, we will further investigate this valuable direction, exploring how to leverage the complementary strengths of medical agents and universal foundation models to advance medical image analysis.
		
	}
	
	\section{Methods}
	\label{sec_method}
	
	\subsection{Details of UniBiomed}
	\label{sec_method_framework}
	
	\begin{figure*}
		\centering
		\includegraphics[width=1\linewidth]{./fig/Extended_framework.pdf}
		\caption{\rebut{The architecture of UniBiomed. Specifically, the user instruction is processed by a pre-trained text tokenizer BERT~\cite{bert} and input to LLM. The input image is encoded by the vision encoder of MLLM, and then the vision tokens are also injected into LLM for generating the diagnostic findings. Following previous methods~\cite{lisa,sa2va,glamm}, we combine the output of the MLLM with the tokenized user instructions as the language embeddings and inject them into SAM2’s prompt encoder for producing segmentation masks. Notably, we use LoRA~\cite{lora} to finetune the MLLM as previous methods~\cite{lisa,sa2va}. In SAM2, the prompt encoder and mask decoder are fine-tuned, while the vision encoder is frozen.}}
		\label{fig_extened_framework}
	\end{figure*}
	
	UniBiomed is based on a novel integration of Multi-modal Large Language Model (MLLM)~\cite{internvl,llava} and the Segment Anything Model (SAM)~\cite{sam,sam2}. Concretely, MLLM is responsible for interpreting multi-modal images and generating text descriptions, while SAM is employed to segment the target biomedical objects based on the given text instructions. 
	
	MLLM consists of a vision encoder~\cite{vit} and a Large Language Model (LLM)~\cite{llama}. The vision encoder is a typical Vision Transformer (ViT)~\cite{vit} model, which is pre-trained by CLIP~\cite{clip} for vision-language alignment~\cite{internvl_original,Qwen-VL,llava}. The LLM~\cite{llama} is responsible for parsing the text instructions from users and generating corresponding answers.
	Given an image and a text instruction, the vision encoder encodes the image into visual tokens, while a text tokenizer~\cite{bert} will process the text into language tokens. These visual and language tokens are then fed into the LLM to generate textual outputs, enabling UniBiomed to tackle diverse text generation tasks.
	
	SAM~\cite{sam,sam2} is further leveraged for tackling the segmentation task, which comprises a vision encoder, a prompt encoder, and a mask decoder. The vision encoder is based on Hiera~\cite{hiera}, a hierarchical ViT~\cite{vit} architecture. 
	The image features extracted by SAM’s vision encoder are passed to the mask decoder, where the prompt encoder encodes the user prompts and prompts the mask encoder to generate segmentation masks. 
	Unlike conventional SAM~\cite{sam,sam2}, which relies on visual prompts (\emph{e.g.}, points or bounding boxes), the SAM in UniBiomed is based on the textual prompts generated by MLLM. 
	\rebut{Following previous methods~\cite{lisa,sa2va,glamm}, UniBiomed combines the output of the MLLM with the tokenized user instructions as the language embeddings and injects them into SAM’s prompt encoder, enabling text-guided segmentation. 
		Specifically, the user instructions are first encoded by the pre-trained text tokenizer (\emph{i.e.}, BERT~\cite{bert}) and forwarded by the LLM. The LLM uses the standard new-token-prediction strategy to generate new tokens following the instructions. Then, we concatenate the embeddings of these generated tokens with the embeddings of the instructions, and inject them into the prompt encoder of SAM for prompting the SAM model.} 
	This effectively bridges the MLLM and SAM for grounded interpretation~\cite{sa2va,lisa}. Notably, following the previous methods~\cite{sa2va,lisa}, we adopt a special token ``[SEG]'' to instruct the mask decoder to produce segmentation masks. While for the tasks that do not require segmentation predictions, \emph{e.g.}, ROI classification and region-aware report generation, this special token is discarded, and the model will not produce segmentation predictions. 
	
	Notably, the success of UniBiomed stems from the complementary integration of MLLM and SAM. \textbf{First}, unlike the original SAM~\cite{sam,sam2}, which relies on visual prompts, UniBiomed employs language embeddings as textual prompts to guide SAM's mask decoder. This innovation eliminates the need for manually crafting precise bounding boxes—a major bottleneck in segmenting biomedical images with dense, irregularly shaped objects~\cite{biomedparse}. \textbf{Second}, segmentation training in UniBiomed implicitly enhances the diagnostic accuracy of the MLLM~\cite{glamm,lisa,sa2va}. Since segmentation predictions depend on the MLLM's textual prompts, the back-propagation process propagates supervision signals from the segmentation masks back to the MLLM, refining its ability to generate precise prompts. This synergistic integration not only improves segmentation performance but also strengthens the MLLM's diagnostic capabilities, creating a mutually reinforcing loop between the two components.
	
	Following the state-of-the-art architecture Sa2VA~\cite{sa2va}, we adopt InternVL2.5~\cite{internvl} as the foundation MLLM and SAM2-hiera-large~\cite{sam2} as the segmentation model. To preserve the learned knowledge of the strong pre-trained MLLM, we leverage LoRA~\cite{lora} to perform efficient fine-tuning of LLM, and freeze the vision encoder of MLLM entirely. For SAM2~\cite{sam2}, following previous settings~\cite{sa2va,lisa}, we freeze the vision encoder and fine-tune the prompt encoder and mask decoder only.
	
	We aggregate different sources of datasets into a universal training process. The overall training loss consists of a text generation loss $L_{text}$ and a segmentation loss $L_{seg}$. The text generation loss $L_{text}$ is an auto-regressive cross-entropy loss as standard LLM~\cite{llava,lisa,sa2va}. The segmentation loss $L_{seg}$ is a combination of per-pixel binary cross-entropy (BCE) loss $L_{BCE}$ and typical dice loss $L_{dice}$. We balance the weights of loss functions, and the overall loss $L$ is defined as:
	\begin{equation}\label{loss_overall}
		L = L_{text} + L_{seg}, ~~L_{seg} = {\lambda}_{BCE}*L_{BCE} + {\lambda}_{dice}*L_{dice}.
	\end{equation}
	Following previous methods~\cite{sa2va,lisa,biomedparse}, the coefficient of loss functions ${\lambda}_{BCE}$ and ${\lambda}_{dice}$ are defined as $2$ and $0.5$, respectively.
	The per-pixel binary cross-entropy (BCE) loss $L_{BCE}$ is defined as:
	\begin{equation}\label{bce_loss}
		L_{BCE} = -\frac{1}{N} \sum_{i=1}^N \left[y_i\log(p_i) + (1-y_i)\log(1-p_i)\right]
	\end{equation}
	where $y_i \in \{0,1\}$ is the ground truth label, $p_i \in [0,1]$ is the predicted probability, and $N$ is the total number of pixels.
	The dice loss $L_{dice}$ is given by:
	\begin{equation}\label{dice_loss}
		L_{dice} = 1 - \frac{2\sum_{i=1}^N p_i y_i + \epsilon}{\sum_{i=1}^N p_i + \sum_{i=1}^N y_i + \epsilon}
	\end{equation}
	where $\epsilon$ is a small constant for numerical stability.
	
	Specifically, for the ROI classification and region-aware report generation tasks, the $L_{seg}$ will be discarded during training. In this case, the loss function $L$ is formulated as:
	\begin{equation}\label{loss_overall_}
		L = L_{text}.
	\end{equation}
	
	\subsection{Implementation}
	\label{sec_method_implementation}
	
	\rebut{
		\textbf{Pre-processing 3D medical images (CT/MRI)}. For pre-processing, we simply follow the settings of BiomedParse~\cite{biomedparse} for fair comparisons, which are also widely adopted in recent medical vision-language models~\cite{medtrinity,lingshu,hulumed}. Specifically, we first conduct normalization on the 3D images, \emph{i.e.}, window-adjustment for CT and Z-score normalization for MRI. In window-adjustment, we set the clipped Hopfield Units (HU) as $[-1000, 500]$ for chest CT and $[-175, 250]$ for abdomen CT. After normalization, we crop the 3D images along the axial axis to get the 2D slices.
	}
	
	For dataset curation, we adopt some readily available toolkits from SA2VA~\cite{sa2va}, RadGenome~\cite{radgenome}, VoCo~\cite{voco,voco-v1}, BiomedParse~\cite{biomedparse}, and MedTrinity~\cite{medtrinity} to transform the datasets into a uniform VQA format. Some examples are shown in \textbf{Extended Data Figure~\ref{fig_extended_vqa_format}}. For fair comparisons, we resize all images to $1024 \times 1024$ sizes for training and validation.
	We use Pytorch~\cite{pytorch} to conduct all training and validation experiments. The training of UniBiomed is conducted on 8 $\times$ NVIDIA H800 (80G) GPUs for 10 epochs, which takes about 5 days to finish the training. 
	All the inference tasks can be established within one NVIDIA 3090 GPU (24GB). \rebut{The seed is set to 42.} We provide the available code links of comparison methods in \textbf{Extended Data Table~\ref{table_code_source}}. 
	More details of training and validation are shown in \textbf{Extended Data Table~\ref{table_preprocess}}. 
	
	\subsection{Evaluation}
	\label{sec_method_evaluation}
	
	Following previous methods~\cite{medsam,biomedparse}, we did not conduct validation during training to select hyperparameters for fair comparisons. \rebut{
		For each internal dataset, we strictly split the dataset into 80\% training and 20\% test sets as the previous method~\cite{biomedparse} for fair comparisons. For 3D CT/MRI datasets, we split the datasets at the volume-level.} To avoid information leakage, all the external datasets are unseen in the training of our method and all comparison methods, and we conduct direct inference on these datasets. 14 external datasets for segmentation are listed in \textbf{Extended Data Table~\ref{table:External_dsc}}. 
	
	\textbf{Comparison methods}. In this work, we focus on evaluating the effectiveness of grounded interpretation. Thus, we compare segmentation foundation models on segmentation tasks and compare medical MLLMs on medical diagnosis tasks.
	
	We first compare MedSAM~\cite{medsam}, BiomedParse~\cite{biomedparse}, SegVol~\cite{segvol}, and SAT~\cite{SAT} in the biomedical image segmentation task. Specifically, MedSAM, BiomedParse, SegVol, and SAT are generalist models that can be applied to multiple datasets. MedSAM and SegVol adopt visual prompts while BiomedParse and SAT use textual prompts for segmentation. SegVol and SAT are only applicable to 3D medical images like CT and MRI, which are not available for multi-modal biomedical images. In addition, the methods based on visual prompts, MedSAM and SegVol, are not applicable to the biomedical objects with irregular shapes~\cite{biomedparse}. 
	Most of the segmentation datasets used in our method have already been trained in these models. \textbf{For fair comparisons, we fine-tune these models on the internal segmentation datasets used in our method}.
	We conduct five experimental runs and report the confidence ranges to ensure reliable results.
	The details of datasets are shown in \textbf{Extended Data Table~\ref{table:dataset}}. 
	The results are shown in \textbf{Extended Data Tables~\ref{table:internal_dsc} and \ref{table:External_dsc}}.
	
	For the text generation task, we first compare GLaMM~\cite{glamm} and LISA~\cite{lisa} in grounded disease recognition and grounded report generation. Specifically, we re-implement these methods on our curated datasets for fair comparisons. The details of datasets for grounded disease recognition are shown in \textbf{Extended Data Table~\ref{table:dataset_grounded_disease}}. We evaluate the performance of grounded report generation on the RadGenome~\cite{radgenome,ctrate} dataset. 
	The results are shown in \textbf{Extended Data Tables~\ref{table:Grounded_disease} and \ref{table:RADGENOME}}.
	
	Then, we compare MedRegA~\cite{MedRegA} and MedPLIB~\cite{MedPLIB} in ROI classification. These two medical MLLMs~\cite{MedRegA,MedPLIB} have been trained on large-scale ROI classification datasets. For fair comparisons, we further fine-tune them on our dataset to ensure the training datasets are consistent. Following MedPLIB~\cite{MedPLIB}, for ROI classification, we degrade the segmentation masks of segmentation datasets to bounding boxes as visual prompts for representing the ROIs, then predict the biomedical classes within the ROIs. Following ViP-LLaVA~\cite{vipllava}, we directly overlay bounding boxes onto the biomedical images, eliminating the need for complex region encodings for indicating ROIs. The datasets for ROI classification are shown in \textbf{Extended Data Table~\ref{table:dataset}}. 
	The results are shown in \textbf{Extended Data Table~\ref{table:ROI}}.
	
	Furthermore, we compare InternVL2.5~\cite{internvl}, LLaVA-Med~\cite{Llava-med}, MedRegA~\cite{MedRegA}, and MedPLIB~\cite{MedPLIB} in region-aware report generation. Notably, the medical MLLMs~\cite{Llava-med,MedRegA,MedPLIB} are trained on the VQA and report generation datasets used in this method, enabling us to conduct fair comparisons. We report the results on the MedTrinity~\cite{medtrinity} dataset. Specifically, the MedTrinity dataset is aggregated from 23 report generation datasets, as shown in \textbf{Extended Data Table~\ref{table:dataset_language}}. In this work, we report the overall results on the whole MedTrinity dataset instead of the separate datasets. 
	The results are shown in \textbf{Extended Data Table~\ref{table:medtrinity}}.

	\rebut{
		\textbf{Discussion with slice selection}. Slice selection methods are proposed for the variants of medical SAM~\cite{ma2025medsam2,medsam2,dong2024segment,sengupta2025sam} that are based on visual prompts (\emph{e.g.}, bounding boxes, points, or scribbles). For example, Dong et al.~\cite{dong2024segment} discussed several slice selection methods, \emph{i.e.}, select which slices for clinicians to annotate. 
		However, our UniBiomed does not rely on visual prompts but only demands text prompts. Given a 3D volume, the same text prompt is used for all of the slices. Thus, our model does not require such slice selection methods~\cite{dong2024segment}, eliminating slice selection bias.
	}
	
	\textbf{Evaluation metrics}. For the evaluation of segmentation, the standard Dice score (\%) is employed to evaluate the performance. Dice score is calculated as:
	\begin{equation}\label{eqn_dsc}
		\text{Dice}(P, G) = \frac{2 |P \cap G|}{|P| + |G|},
	\end{equation}
	where $P$ denotes the segmentation predictions, $G$ is the ground truth of segmentation labels. 
	
	For the task of grounded disease recognition, we present the results of both dice scores and accuracy. The accuracy is calculated as:
	\begin{equation}\label{eqn_accuracy}
		\text{Accuracy} = \frac{1}{N} \sum_{i=1}^{N} 1(y_i = \hat{y}_i),
	\end{equation}
	where $N$ is the total number of samples. $y$ and $\hat{y}$ are the predictions and labels, respectively. $1(y_i = \hat{y}_i)$ denotes that the prediction of $i_{th}$ sample is correct.
	
	For the task of ROI classification, we simply use the accuracy metric to measure the performance. As for the region-aware report generation task, we further employ Bilingual Evaluation Understudy (BLEU)~\cite{bleu}, Metric for Evaluation of Translation with Explicit ORdering (METEOR)~\cite{meteor}, and Recall-Oriented Understudy for Gisting Evaluation with Longest Common Subsequence (ROUGE\_L)~\cite{rouge} for evaluation.
	
	Specifically, the formulation of BLEU is as follows:
	\begin{equation}\label{eqn_bleu}
		\text{BLEU} = BP \cdot \exp\left(\sum_{n=1}^N w_n \log p_n\right),
	\end{equation}
	where BP is the brevity penalty, $p_n$ are modified n-gram precisions, and $w_n$ are weights. The BLEU variants, \emph{i.e.}, BLEU1, BLEU2, BLEU3, and BLEU4 are differ in their n-gram scope:
	\begin{equation}\label{eqn_bleu1}
		\text{BLEU-1} = BP \cdot \exp(\log p_1),
	\end{equation}
	
	\begin{equation}\label{eqn_bleu2}
		\text{BLEU-2} = BP \cdot \exp\left(\frac{1}{2} \sum_{n=1}^2 \log p_n\right),
	\end{equation}
	
	\begin{equation}\label{eqn_bleu3}
		\text{BLEU-3} = BP \cdot \exp\left(\frac{1}{3} \sum_{n=1}^3 \log p_n\right),
	\end{equation}
	
	\begin{equation}\label{eqn_bleu4}
		\text{BLEU-4} = BP \cdot \exp\left(\frac{1}{4} \sum_{n=1}^4 \log p_n\right).
	\end{equation}
	
	The METEOR~\cite{meteor} is an automatic evaluation metric for evaluating the quality of machine translation, which not only considers vocabulary matching but also combines word order similarity and alignment information. It is calculated as:
	\begin{equation}\label{eqn_meteor}
		\text{METEOR} = \frac{1}{m} \cdot \sum_{g \in \text{gold}} \max_{h \in \text{hyp}} \text{Precision}(g, h),
	\end{equation}
	where $m$ is the number of gold standard (reference) sentences, and $Precision(g, h)$ refers to the precision score between a specific gold standard sentence (g) and a hypothesis
	sentence (h) from the set of all gold standard sentences (gold) and the set of all hypothesis sentences (hyp).
	
	ROUGE-L is an automatic evaluation metric for assessing machine-generated text by measuring the longest common subsequence (LCS) between the hypothesis and reference. It computes recall, precision, and their harmonic mean (F-score) as follows:
	
	\begin{equation}
		R_{\text{LCS}} = \frac{\text{LCS}(X, Y)}{|Y|}, \quad
		P_{\text{LCS}} = \frac{\text{LCS}(X, Y)}{|X|}, \quad
		F_{\text{LCS}} = \frac{(1 + \beta^2) R_{\text{LCS}} P_{\text{LCS}}}{R_{\text{LCS}} + \beta^2 P_{\text{LCS}}},
	\end{equation}
	
	where $X$ is the hypothesis, $Y$ is the reference, and 
	$\beta$ controls recall emphasis. For multiple references, it averages the best $F_{\text{LCS}}$ across pairs, similar to METEOR in Equation~\ref{eqn_meteor}. Unlike METEOR, it ignores synonyms and alignment, focusing on word-order-agnostic overlap.
	
	For the grounded report generation evaluation, we also use (BLEU)~\cite{bleu}, METEOR~\cite{meteor}, and ROUGE\_L~\cite{rouge} as the metrics. In addition, we further use dice scores to measure the performance of segmentation.
	
	\subsection{Dataset descriptions}
	
	\rebut{
		This study involves 10 diverse imaging modalities. The number of images for each modality is as follows: X-ray (79,212), CT (8,171,448), MRI (6,630,101), Pathology (4,029,165), Ultrasound (26,066), Fundus (1,436), Dermoscopy (6,616), Endoscope (4,061), OCT (1,484), and PET (11,561). These images are collected from 84 diverse datasets, as shown in \textbf{Extended Data Table~\ref{table_dataset_descriptions}}.
		Specifically, the 46 internal datasets and 14 external datasets for segmentation, ROI classification, and disease recognition are listed in \textbf{Extended Data Tables~\ref{table:internal_dsc} and \ref{table:External_dsc}}, respectively. The datasets for report generation are based on the Medtrinity~\cite{medtrinity} and RadGenome~\cite{radgenome} datasets. The aggregated 23 datasets in Medtrinity~\cite{medtrinity} are listed in \textbf{Extended Data Table~\ref{table:dataset_language}}. For the internal and external data split, we split 14 external datasets with the condition that their prediction targets fall within the intended scope of the internal data.
		
		To ensure the quality of our training and validation datasets: (1) All data are sourced from peer-reviewed public datasets that are widely adopted in the community, and all labels are annotated by human experts to guarantee their quality. (2) We avoided using LLMs to generate or process any text descriptions, thus preventing potential LLM hallucinations.
		
		For segmentation, disease recognition, and ROI classification datasets, we derive the text descriptions directly from the expert-annotated labels. For example, in a liver tumor dataset, if radiologists have annotated a liver tumor in an image, we assign the text description ``liver tumor" to it, following established practices in prior works~\cite{biomedparse,SAT,Biomedgpt,Llava-med,medtrinity}. To align the image regions with the text descriptions, we use the human-annotated segmentation masks to match the image regions and texts, ensuring high-quality alignments. 
		For VQA and report generation tasks, the text descriptions are identical to those in the original public datasets, all of which were expert-annotated and underwent rigorous peer review and quality control.
		
		To ensure label consistency across different segmentation datasets, we follow the protocol of BiomedParse~\cite{biomedparse}  for fair comparisons. For example, for liver tumor datasets, we consistently use the label ``liver tumor" and avoid inconsistent terms such as ``liver cancer" or ``liver lesion".
		
		We follow the standard of previous works~\cite{biomedparse,Llava-med,Biomedgpt,lingshu,hulumed,gsco,segvol,medsam,SAT} to aggregate different datasets for training. Following these methods, we have proactively addressed potential biases by aggregating datasets from a wide range of institutions, imaging protocols, and patient demographics. This diversity ensures that our model is exposed to a broad spectrum of clinical scenarios, significantly mitigating the risk of bias from any single source and enhancing its generalizability.
		
		Inter-annotator discrepancy is an inherent challenge when aggregating large-scale datasets, as it is impractical for a single annotator to label all the data. To mitigate its influence, we followed established practices from previous works~\cite{biomedparse,Llava-med,Biomedgpt,lingshu,hulumed,gsco,segvol,medsam,SAT} by exclusively using peer-reviewed public datasets. These datasets have undergone rigorous quality control procedures, which serve to minimize such discrepancies. 
		While this study does not aim to comprehensively analyze dataset biases, we recognize its significance and aim to investigate more advanced bias quantification and mitigation techniques in the future.
		
	}
	
	\section{Data availability}\label{sec_data_avail}
	
	This study incorporates a total of 84 datasets across 10 diverse biomedical imaging modalities. All these datasets are publicly available for research. The details about the data used in this project are listed in \textbf{Extended Data Tables~\ref{table:dataset}, \ref{table:dataset_language}, \ref{table_dataset_descriptions}, and \ref{table_dataset_descriptions_modality}}.
	
	\section{Code availability}\label{sec6}
	
	The codes, datasets, and models of UniBiomed is available at GitHub
	(\href{https://github.com/Luffy03/UniBiomed}{https://github.com/Luffy03/UniBiomed}).
	
	\section{Author contributions}\label{sec8}
	
	
	L.W. designed the framework and conducted the experiments. Y.N., S.H., J.Z., and L.L. contributed to the data acquisition and provided suggestions on the framework. \rebut{T.L., Z.X., and D.C. contributed to the reader studies.
		Y.Z., N.M., V.V., and R.C.K.C. contributed to the evaluation and analyzed the results of UniBiomed.} 
	Y.P. and P.R. polished the manuscript and provided suggestions on the experiments. 
	H.C. and L.W. conceived this work. H.C. supervised the project. All authors discussed the results and contributed to the final manuscript.
	
	\section*{Declaration}
	
	The authors have no conflicts of interest to declare.
	
	\section*{Ethics declaration}\label{sec7}
	
	This project has been reviewed and approved by the Human and Artefacts Research Ethics Committee (HAREC). The protocol number is HREP-2025-0188.
	
	\section*{Acknowledgements}
	This work was supported by the Hong Kong Innovation and Technology Commission (Project No. MHP/002/22, GHP/006/22GD and ITCPD/17-9), HKUST (Project No. FS111), and the Research Grants Council of the Hong Kong Special Administrative Region, China (Project Reference Number: T45-401/22-N). 
	We also thank the support of HKUST SuperPOD for providing the GPU platform for model training.

	
	\bibliography{sn-bibliography}

	\newpage
	\begin{appendices}
		
		\section{Extended Data}\label{secA1}
		
		\renewcommand{\arraystretch}{1.6}
		\begin{table*}[!h]
			\setlength{\abovecaptionskip}{0.pt}
			\setlength{\belowcaptionskip}{-0.em}
			\centering
			\footnotesize
			\caption{\textbf{Comparison of different representative biomedical models}. Compared with previous methods, UniBiomed supports various modalities, prompts, and tasks. \emph{LLM} denotes Large-Language Models. \emph{Region-Aware Diagnosis} denotes models that can interpret regions of interest defined by users via visual prompts. \emph{Pixel-level Grounding} indicates models that can generate text descriptions with corresponding segmentation masks. \emph{End-to-End Training} indicates the models are trained in an end-to-end process.}
			\begin{threeparttable}
				\resizebox{1.0\textwidth}{!}{
					\begin{tabular}{l|c|cc|ccc|c|c|c|c}
						\toprule[1.2pt]
						\multirow{2}{*}{\textbf{Method}} &\textbf{Multi-Modal} &\multicolumn{2}{c|}{\textbf{Support Prompts}} &\multicolumn{3}{c|}{\textbf{Support Response}}  &\multirow{2}{*}{\textbf{LLM}} &\textbf{Region-Aware}  &\textbf{Pixel-level} &\textbf{End-End}
						\\
						&\textbf{Images} &\textbf{Visual} &\textbf{Text} &\textbf{Mask} &\textbf{Text} &\textbf{Mask+Text} 
						&
						&\textbf{Diagnosis} &\textbf{Grounding} &\textbf{Training}\\
						\hline
						MedSAM~\cite{medsam} 
						&\textcolor{dark-green}{\CheckmarkBold}
						&\textcolor{dark-green}{\CheckmarkBold} &\textcolor{red}{\XSolidBrush} 
						&\textcolor{dark-green}{\CheckmarkBold} &\textcolor{red}{\XSolidBrush} 
						&\textcolor{red}{\XSolidBrush}
						&\textcolor{red}{\XSolidBrush}
						&\textcolor{red}{\XSolidBrush} 
						&\textcolor{red}{\XSolidBrush} 
						&\textcolor{dark-green}{\CheckmarkBold}\\
						
						BiomedParse~\cite{biomedparse} 
						&\textcolor{dark-green}{\CheckmarkBold} 
						&\textcolor{red}{\XSolidBrush}  
						&\textcolor{dark-green}{\CheckmarkBold} 
						&\textcolor{dark-green}{\CheckmarkBold} &\textcolor{red}{\XSolidBrush} 
						&\textcolor{red}{\XSolidBrush}
						&\textcolor{red}{\XSolidBrush} 
						&\textcolor{red}{\XSolidBrush} 
						&\textcolor{red}{\XSolidBrush}
						&\textcolor{dark-green}{\CheckmarkBold}\\
						
						3DSAM~\cite{3dsam} 
						&\textcolor{red}{\XSolidBrush} 
						&\textcolor{dark-green}{\CheckmarkBold} &\textcolor{red}{\XSolidBrush} 
						&\textcolor{dark-green}{\CheckmarkBold}  &\textcolor{red}{\XSolidBrush} 
						&\textcolor{red}{\XSolidBrush}
						&\textcolor{red}{\XSolidBrush} 
						&\textcolor{red}{\XSolidBrush} 
						&\textcolor{red}{\XSolidBrush}
						&\textcolor{red}{\XSolidBrush}\\
						
						SegVol~\cite{segvol} 
						&\textcolor{red}{\XSolidBrush} 
						&\textcolor{dark-green}{\CheckmarkBold} &\textcolor{red}{\XSolidBrush}
						&\textcolor{dark-green}{\CheckmarkBold}  &\textcolor{red}{\XSolidBrush} 
						&\textcolor{red}{\XSolidBrush}
						&\textcolor{red}{\XSolidBrush} 
						&\textcolor{red}{\XSolidBrush} 
						&\textcolor{red}{\XSolidBrush}
						&\textcolor{dark-green}{\CheckmarkBold}\\
						
						SAT~\cite{SAT} 
						&\textcolor{red}{\XSolidBrush} 
						&\textcolor{red}{\XSolidBrush} 
						&\textcolor{dark-green}{\CheckmarkBold} 
						&\textcolor{dark-green}{\CheckmarkBold}  &\textcolor{red}{\XSolidBrush} 
						&\textcolor{red}{\XSolidBrush}
						&\textcolor{red}{\XSolidBrush}
						&\textcolor{red}{\XSolidBrush} 
						&\textcolor{red}{\XSolidBrush}
						&\textcolor{dark-green}{\CheckmarkBold}\\
						
						MedSAM2~\cite{medsam2} 
						&\textcolor{dark-green}{\CheckmarkBold} 
						&\textcolor{dark-green}{\CheckmarkBold} &\textcolor{red}{\XSolidBrush} 
						&\textcolor{dark-green}{\CheckmarkBold} &\textcolor{red}{\XSolidBrush} 
						&\textcolor{red}{\XSolidBrush}
						&\textcolor{red}{\XSolidBrush} 
						&\textcolor{red}{\XSolidBrush} 
						&\textcolor{red}{\XSolidBrush}
						&\textcolor{dark-green}{\CheckmarkBold}\\
						
						M3D~\cite{m3d} 
						&\textcolor{red}{\XSolidBrush} 
						&\textcolor{red}{\XSolidBrush}  
						&\textcolor{dark-green}{\CheckmarkBold} 
						&\textcolor{dark-green}{\CheckmarkBold}
						&\textcolor{dark-green}{\CheckmarkBold} 
						&\textcolor{red}{\XSolidBrush}
						&\textcolor{dark-green}{\CheckmarkBold}
						&\textcolor{red}{\XSolidBrush} 
						&\textcolor{red}{\XSolidBrush}
						&\textcolor{red}{\XSolidBrush}\\
						
						RadFM~\cite{RADFM} 
						&\textcolor{red}{\XSolidBrush} 
						&\textcolor{red}{\XSolidBrush}  
						&\textcolor{dark-green}{\CheckmarkBold} 
						&\textcolor{red}{\XSolidBrush} 
						&\textcolor{dark-green}{\CheckmarkBold} 
						&\textcolor{red}{\XSolidBrush}
						&\textcolor{dark-green}{\CheckmarkBold}
						&\textcolor{red}{\XSolidBrush} 
						&\textcolor{red}{\XSolidBrush}
						&\textcolor{red}{\XSolidBrush}\\
						
						LLaVA-Med~\cite{Llava-med} 
						&\textcolor{dark-green}{\CheckmarkBold} 
						&\textcolor{red}{\XSolidBrush}  
						&\textcolor{dark-green}{\CheckmarkBold} 
						&\textcolor{red}{\XSolidBrush} 
						&\textcolor{dark-green}{\CheckmarkBold} 
						&\textcolor{red}{\XSolidBrush}
						&\textcolor{dark-green}{\CheckmarkBold}
						&\textcolor{red}{\XSolidBrush} 
						&\textcolor{red}{\XSolidBrush}
						&\textcolor{red}{\XSolidBrush}\\
						
						BiomedGPT~\cite{Biomedgpt} 
						&\textcolor{dark-green}{\CheckmarkBold} 
						&\textcolor{red}{\XSolidBrush}  
						&\textcolor{dark-green}{\CheckmarkBold} 
						&\textcolor{red}{\XSolidBrush} 
						&\textcolor{dark-green}{\CheckmarkBold} 
						&\textcolor{red}{\XSolidBrush}
						&\textcolor{dark-green}{\CheckmarkBold}
						&\textcolor{red}{\XSolidBrush} 
						&\textcolor{red}{\XSolidBrush}
						&\textcolor{dark-green}{\CheckmarkBold}\\
						
						fVLM~\cite{fvlm} 
						&\textcolor{red}{\XSolidBrush} 
						&\textcolor{red}{\XSolidBrush}  
						&\textcolor{dark-green}{\CheckmarkBold} 
						&\textcolor{red}{\XSolidBrush} 
						&\textcolor{dark-green}{\CheckmarkBold} 
						&\textcolor{red}{\XSolidBrush}
						&\textcolor{red}{\XSolidBrush}
						&\textcolor{dark-green}{\CheckmarkBold}
						&\textcolor{red}{\XSolidBrush}
						&\textcolor{dark-green}{\CheckmarkBold}\\
						
						MedRegA~\cite{MedRegA} 
						&\textcolor{dark-green}{\CheckmarkBold} 
						&\textcolor{dark-green}{\CheckmarkBold}  
						&\textcolor{dark-green}{\CheckmarkBold} 
						&\textcolor{red}{\XSolidBrush} 
						&\textcolor{dark-green}{\CheckmarkBold} 
						&\textcolor{red}{\XSolidBrush}
						&\textcolor{dark-green}{\CheckmarkBold}
						&\textcolor{dark-green}{\CheckmarkBold} 
						&\textcolor{red}{\XSolidBrush}
						&\textcolor{dark-green}{\CheckmarkBold}\\
						
						MedPLIB~\cite{MedPLIB} &\textcolor{dark-green}{\CheckmarkBold} 
						&\textcolor{dark-green}{\CheckmarkBold} &\textcolor{dark-green}{\CheckmarkBold} 
						&\textcolor{dark-green}{\CheckmarkBold}  
						&\textcolor{dark-green}{\CheckmarkBold} 
						&\textcolor{red}{\XSolidBrush} 
						&\textcolor{dark-green}{\CheckmarkBold}
						&\textcolor{dark-green}{\CheckmarkBold} 
						&\textcolor{dark-green}{\CheckmarkBold} 
						&\textcolor{red}{\XSolidBrush}\\
						
						\hline
						\rowcolor{mygray}
						\textbf{UniBiomed} &\textcolor{dark-green}{\CheckmarkBold}
						&\textcolor{dark-green}{\CheckmarkBold} &\textcolor{dark-green}{\CheckmarkBold} 
						&\textcolor{dark-green}{\CheckmarkBold}  
						&\textcolor{dark-green}{\CheckmarkBold} 
						&\textcolor{dark-green}{\CheckmarkBold} 
						&\textcolor{dark-green}{\CheckmarkBold} 
						&\textcolor{dark-green}{\CheckmarkBold} 
						&\textcolor{dark-green}{\CheckmarkBold}
						&\textcolor{dark-green}{\CheckmarkBold}
						\\
						
						\toprule[1.2pt]
					\end{tabular}
				}
			\end{threeparttable}        
			\label{table_novelty}
		\end{table*}
		
		\newpage
		
		\begin{figure*}[!h]
			\centering
			\includegraphics[width=1\linewidth]{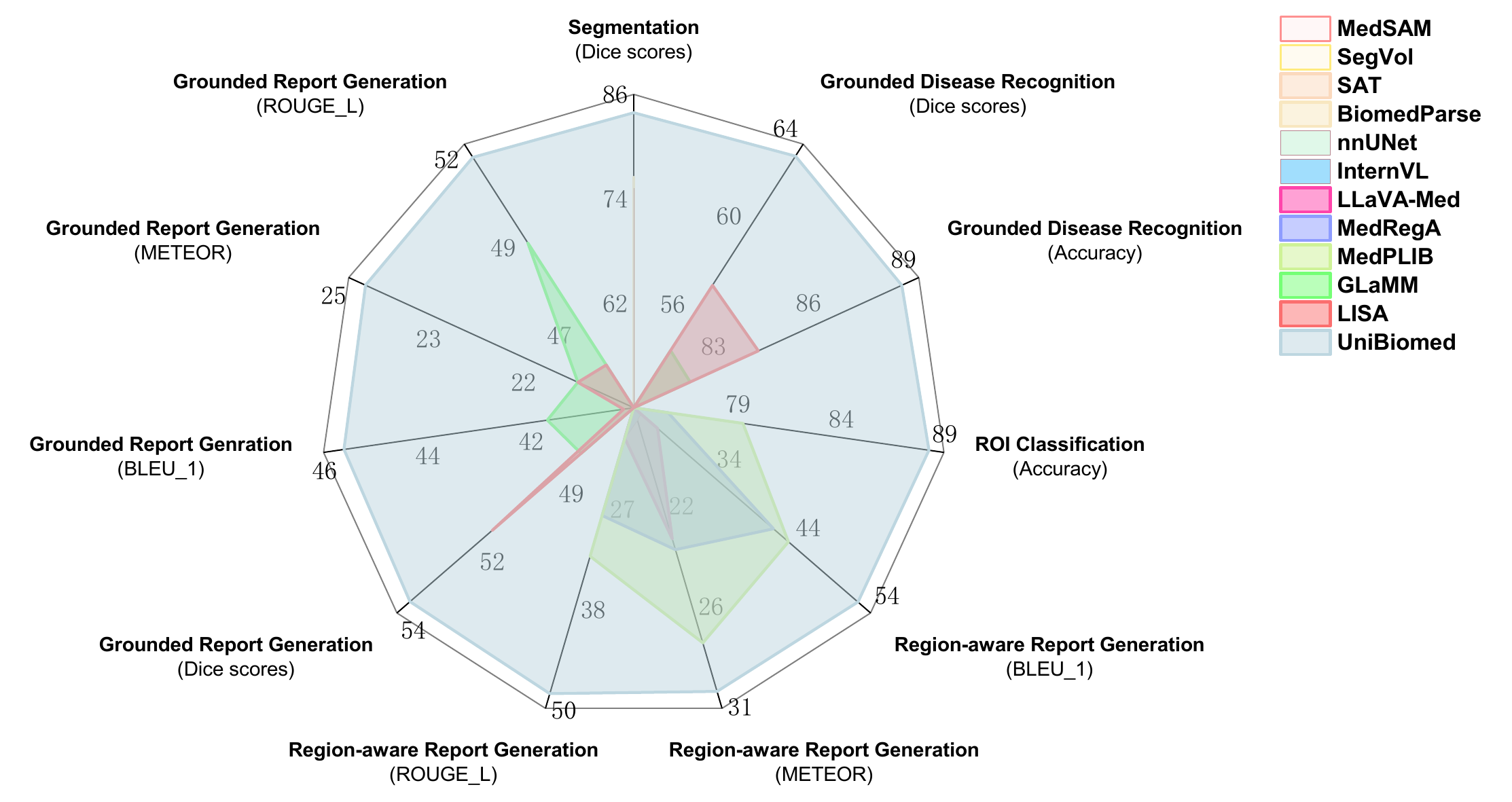}
			\caption{Overall comparisons on various biomedical tasks, including segmentation, disease recognition, ROI classification, region-aware report generation, and grounded report generation. We compare UniBiomed with MedSAM~\cite{medsam}, BiomedParse~\cite{biomedparse}, SegVol~\cite{segvol}, SAT~\cite{SAT}, InternVL2.5~\cite{internvl}, LLaVA-Med~\cite{Llava-med}, MedRegA~\cite{MedRegA}, MedPLIB~\cite{MedPLIB}, GLaMM~\cite{glamm}, and LISA~\cite{lisa}. Notably, while previous methods can address only a limited number of these tasks, UniBiomed excels by delivering state-of-the-art performance across all of them.}
			\label{fig_extended_RADAR}
		\end{figure*}
		
		\clearpage
		
		\begin{table*}[!h]
			\renewcommand{\arraystretch}{1.4} 
			\center
			\caption{\textbf{61 datasets across 10 modalities for segmentation, ROI classification, and grounded disease recognition}. 
				47 and 14 of them are used for internal and external validation, respectively. The external datasets are annotated with *.
			}
			\label{table:dataset}
			\resizebox{1.0\textwidth}{!}{
				\begin{tabular}{lll}
					\toprule
					\rowcolor{lightgray} \textbf{Dataset} &\textbf{Modality} & \textbf{Link} \\
					\midrule
					CXRMask~\cite{CXRMask} &X-Ray &\href{https://datasetninja.com/chest-xray}{https://datasetninja.com/chest-xray}\\
					
					Radiography~\cite{chowdhury2020can} &X-Ray &\href{https://www.kaggle.com/datasets/tawsifurrahman/covid19-radiography-database}{https://www.kaggle.com/datasets/tawsifurrahman/covid19-radiography-database}\\
					
					COVID-QU-Ex~\cite{COVID-QU-Ex} &X-Ray &\href{https://www.kaggle.com/datasets/anasmohammedtahir/covidqu}{https://www.kaggle.com/datasets/anasmohammedtahir/covidqu}\\
					
					CDD-CESM~\cite{CDD-CESM} &X-Ray &\href{https://www.cancerimagingarchive.net/collection/cdd-cesm/}{https://www.cancerimagingarchive.net/collection/cdd-cesm/}\\
					
					SIIM~\cite{SIIM} &X-Ray &\href{https://www.kaggle.com/datasets/vbookshelf/pneumothorax-chest-xray-images-and-masks}{https://www.kaggle.com/datasets/vbookshelf/pneumothorax-chest-xray-images-and-masks}\\
					
					PanNuke~\cite{pannuke} &Pathology &\href{https://jgamper.github.io/PanNukeDataset/}{https://jgamper.github.io/PanNukeDataset/}\\
					
					GlaS~\cite{glas} &Pathology &\href{https://warwick.ac.uk/fac/cross_fac/tia/data/glascontest/}{https://warwick.ac.uk/fac/cross\_fac/tia/data/glascontest/}\\
					
					CoCaHis~\cite{cocahis} &Pathology &\href{https://cocahis.irb.hr/}{https://cocahis.irb.hr/}\\
					
					CryoNuSeg~\cite{cryonuseg} &Pathology &\href{https://github.com/masih4/CryoNuSeg}{https://github.com/masih4/CryoNuSeg}\\
					
					DigestPath~\cite{DigestPath} &Pathology &\href{https://digestpath2019.grand-challenge.org/}{https://digestpath2019.grand-challenge.org/}\\
					
					SICAPv2~\cite{SICAPv2} &Pathology &\href{https://data.mendeley.com/datasets/9xxm58dvs3/1}{https://data.mendeley.com/datasets/9xxm58dvs3/1}\\
					
					WSSS4LUAD~\cite{WSSS4LUAD} &Pathology &\href{https://wsss4luad.grand-challenge.org/}{https://wsss4luad.grand-challenge.org/}\\
					
					CRAG*~\cite{CRAG} &Pathology &\href{https://warwick.ac.uk/fac/cross\_fac/tia/data/mildnet/}{https://warwick.ac.uk/fac/cross\_fac/tia/data/mildnet/}\\
					
					MoNuSeg*~\cite{MoNuSeg} &Pathology &\href{https://monuseg.grand-challenge.org/}{https://monuseg.grand-challenge.org/}\\
					
					US~\cite{US,CAMUS,FH-PS-AOP} &Ultrasound &\href{https://www.kaggle.com/datasets/ignaciorlando/ussimandsegm}{https://www.kaggle.com/datasets/ignaciorlando/ussimandsegm}\\
					
					REFUGE~\cite{refuge} &Fundus &\href{https://bitbucket.org/woalsdnd/refuge/src}{https://bitbucket.org/woalsdnd/refuge/src}\\
					
					DRIVE~\cite{DRIVE} &Fundus &\href{https://drive.grand-challenge.org/}{https://drive.grand-challenge.org/}\\
					
					UWater~\cite{UWaterlooSkinCancer} &Dermoscopy &\href{https://uwaterloo.ca/}{https://uwaterloo.ca/}\\
					
					NeoPolyp~\cite{NeoPolyp} &Endoscope &\href{https://www.kaggle.com/c/bkai-igh-neopolyp/data}{https://www.kaggle.com/c/bkai-igh-neopolyp/data}\\
					
					OCT-CME~\cite{OCT-CME} &OCT &\href{https://www.kaggle.com/datasets/zeeshanahmed13/intraretinal-cystoid-fluid}{https://www.kaggle.com/datasets/zeeshanahmed13/intraretinal-cystoid-fluid}\\
					
					MSD~\cite{msd} & CT$\&$MRI &\href{http://medicaldecathlon.com/}{http://medicaldecathlon.com/}\\
					
					AMOS~\cite{amos} & CT$\&$MRI &\href{https://amos22.grand-challenge.org/}{https://amos22.grand-challenge.org/}\\
					
					BTCV~\cite{btcv} &CT &\href{https://www.synapse.org/\#!Synapse:syn3193805/wiki/}{https://www.synapse.org/\#!Synapse:syn3193805/wiki}\\
					
					WORD~\cite{word} & CT &\href{https://github.com/HiLab-git/WORD}{https://github.com/HiLab-git/WORD}\\
					
					FLARE22~\cite{FLARE}  &CT &\href{https://flare22.grand-challenge.org/}{https://flare22.grand-challenge.org/}\\
					
					FLARE23*~\cite{FLARE}  &CT &\href{https://codalab.lisn.upsaclay.fr/competitions/12239}{https://codalab.lisn.upsaclay.fr/competitions/12239}\\
					
					
					KiTS23~\cite{KiTS} & CT &\href{https://kits-challenge.org/kits23/}{https://kits-challenge.org/kits23/}\\
					
					AbdomenCT1K*~\cite{AbdomenCT-1K} &CT &\href{https://github.com/JunMa11/AbdomenCT-1K}{https://github.com/JunMa11/AbdomenCT-1K}\\
					
					LIDC-IDRI~\cite{LIDC} &CT &\href{https://www.cancerimagingarchive.net/collection/lidc-idri/}{https://www.cancerimagingarchive.net/collection/lidc-idri/}\\
					
					AIIB~\cite{AIIB} &CT &\href{https://codalab.lisn.upsaclay.fr/competitions/13238}{https://codalab.lisn.upsaclay.fr/competitions/13238}\\
					
					COVID-CT~\cite{COVID} &CT &\href{https://covid-segmentation.grand-challenge.org/}{https://covid-segmentation.grand-challenge.org/}\\
					
					AVT~\cite{AVT} &CT &\href{https://figshare.com/articles/dataset/Aortic_Vessel_Tree_AVT_CTA_Datasets_and_Segmentations/14806362}{https://figshare.com/articles/dataset/Aortic\_Vessel\_Tree\_AVT\_CTA\_Datasets\_and\_Segmentations/}\\
					
					CHAOS*~\cite{CHAOS} & CT &\href{https://chaos.grand-challenge.org/Combined_Healthy_Abdominal_Organ_Segmentation/}{https://chaos.grand-challenge.org/}\\
					
					IRCADb*~\cite{3D-IRCADb}  &CT &\href{https://www.ircad.fr/research/data-sets/liver-segmentation-3d-ircadb-01}{https://www.ircad.fr/research/data-sets/liver-segmentation-3d-ircadb-01}\\
					
					SLIVER07*~\cite{SLIVER07} &CT &\href{https://sliver07.grand-challenge.org/}{https://sliver07.grand-challenge.org/}\\
					
					HCCTACE*~\cite{HCCTACE} &CT &\href{https://www.cancerimagingarchive.net/collection/hcc-tace-seg}{https://www.cancerimagingarchive.net/collection/hcc-tace-seg}\\
					
					TCIAPancreas*~\cite{tcia-panc} &CT &\href{https://www.cancerimagingarchive.net/collection/pancreas-ct/}{https://www.cancerimagingarchive.net/collection/pancreas-ct/}\\
					
					QUBIQ*~\cite{Qubiq} &CT &\href{https://qubiq21.grand-challenge.org/QUBIQ/}{https://qubiq21.grand-challenge.org/QUBIQ/}\\
					
					Rider*~\cite{RIDER-LungCT-Seg} &CT &\href{https://www.cancerimagingarchive.net/analysis-result/rider-lungct-seg/}{https://www.cancerimagingarchive.net/analysis-result/rider-lungct-seg/}\\
					
					PANORAMA*~\cite{PANORAMA} &CT &\href{https://panorama.grand-challenge.org/}{https://panorama.grand-challenge.org/}\\
					
					AbdomenAtlas*~\cite{atlas,bassi2024touchstone} &CT &\href{https://github.com/MrGiovanni/AbdomenAtlas}{https://github.com/MrGiovanni/AbdomenAtlas}\\
					
					RadGenome~\cite{radgenome} &CT &\href{https://huggingface.co/datasets/RadGenome/RadGenome-ChestCT}{https://huggingface.co/datasets/RadGenome/RadGenome-ChestCT}\\
					
					LGG~\cite{LGG}  &MRI &\href{https://www.kaggle.com/datasets/mateuszbuda/lgg-mri-segmentation}{https://www.kaggle.com/datasets/mateuszbuda/lgg-mri-segmentation}\\
					
					ACDC~\cite{ACDC}  &MRI &\href{https://www.creatis.insa-lyon.fr/Challenge/acdc/databases.html}{https://www.creatis.insa-lyon.fr/Challenge/acdc/databases.html}\\
					
					BraTS21*~\cite{brats} &MRI &\href{https://www.synapse.org/Synapse:syn51156910/wiki/621282}{https://www.synapse.org/Synapse:syn51156910/wiki/621282}\\
					
					AutoPET~\cite{autopet} & PET & \href{https://autopet.grand-challenge.org/}{https://autopet.grand-challenge.org/} \\
					
					\bottomrule
				\end{tabular}
			}
		\end{table*}
		
		\newpage
		
		\begin{table*}[!tbh]
			\renewcommand{\arraystretch}{1.6} 
			\center
			\caption{\textbf{23 Datasets across 10 modalities for training and validation of report generation}. Multi-modal contains: X-ray, Pathology, Endoscopy, Ultrasound, Dermoscopy, Fundus, PET, CT, OCT, and MRI. The bounding box annotations are generated by MedTrinity~\cite{medtrinity}.}
			\label{table:dataset_language}
			\resizebox{1.0\textwidth}{!}{
				\begin{tabular}{lll}
					\toprule
					\rowcolor{lightgray} \textbf{Dataset} & \textbf{Modality} & \textbf{Link} \\
					\midrule
					NIH-CXR~\cite{NIH-CXR} & X-Ray & \href{https://www.kaggle.com/datasets/nih-chest-xrays/data}{https://www.kaggle.com/datasets/nih-chest-xrays/data} \\
					
					Breast Histopathology & Pathology & \href{https://www.kaggle.com/datasets/paultimothymooney/breast-histopathology-images}{https://www.kaggle.com/datasets/paultimothymooney/breast-histopathology-images} \\
					
					BreastCancer & Pathology & \href{https://zenodo.org/records/6633721}{https://zenodo.org/records/6633721} \\
					
					CISC & Pathology & \href{https://academictorrents.com/details/99f2c7b57b95500711e33f2ee4d14c9fd7c7366c}{https://academictorrents.com/details/99f2c7b57b95500711e33f2ee4d14c9fd7c7366c} \\
					
					CPD & Pathology & \href{https://zenodo.org/records/7282326}{https://zenodo.org/records/7282326} \\
					
					IHC4BC~\cite{IHC4BC} & Pathology & \href{https://www.kaggle.com/datasets/akbarnejad1991/ihc4bc-compressed/data}{https://www.kaggle.com/datasets/akbarnejad1991/ihc4bc-compressed/data} \\
					
					NCT-CRC-HE-100K~\cite{NCT-CRC-HE-100K} & Pathology & \href{https://www.kaggle.com/datasets/imrankhan77/nct-crc-he-100k}{https://www.kaggle.com/datasets/imrankhan77/nct-crc-he-100k} \\
					
					PatchGastricADC22~\cite{PatchGastricADC22} & Pathology & \href{https://zenodo.org/records/6550925}{https://zenodo.org/records/6550925} \\
					
					Path-VQA~\cite{pathvqa} & Pathology & \href{https://huggingface.co/datasets/flaviagiammarino/path-vqa}{https://huggingface.co/datasets/flaviagiammarino/path-vqa} \\
					
					TCGA-UT & Pathology & \href{https://zenodo.org/records/5889558}{https://zenodo.org/records/5889558} \\
					
					PTCGA~\cite{PTCGA} & Pathology & \href{https://drive.google.com/drive/folders/18CmL-WLyppK1Rk29CgV7ib5MACFzg5ei}{https://drive.google.com/drive/folders/18CmL-WLyppK1Rk29CgV7ib5MACFzg5ei} \\
					
					BHX & CT & \href{https://physionet.org/content/bhx-brain-bounding-box/1.1/}{https://physionet.org/content/bhx-brain-bounding-box/1.1/} \\
					
					DeepLesion~\cite{DeepLesion} & CT & \href{https://huggingface.co/datasets/farrell236/DeepLesion}{https://huggingface.co/datasets/farrell236/DeepLesion} \\
					
					CT-RATE~\cite{ctrate} & CT & \href{https://huggingface.co/datasets/ibrahimhamamci/CT-RATE}{https://huggingface.co/datasets/ibrahimhamamci/CT-RATE} \\
					
					RadGenome~\cite{radgenome} & CT & \href{https://huggingface.co/datasets/RadGenome/RadGenome-ChestCT}{https://huggingface.co/datasets/RadGenome/RadGenome-ChestCT} \\
					
					BRATS24-MICCAI & MRI & \href{https://www.synapse.org/Synapse:syn53708126}{https://www.synapse.org/Synapse:syn53708126} \\
					
					LLD-MMRI~\cite{LLD-MMRI} & MRI & \href{https://github.com/LMMMEng/LLD-MMRI-Dataset}{https://github.com/LMMMEng/LLD-MMRI-Dataset} \\
					
					MAMA-MIA~\cite{MAMA-MIA} & MRI & \href{https://www.synapse.org/Synapse:syn60868042/wiki/628716}{https://www.synapse.org/Synapse:syn60868042/wiki/628716} \\
					
					VQA-RAD~\cite{VQA-RAD} & X-Ray\&MRI & \href{https://osf.io/89kps/}{https://osf.io/89kps/} \\
					
					SLAKE~\cite{slake} & X-Ray\&CT\&MRI & \href{https://www.med-vqa.com/slake/}{https://www.med-vqa.com/slake/} \\
					
					PMC-OA~\cite{pmc-oa} & Multi-modal & \href{https://huggingface.co/datasets/axiong/pmc_oa}{https://huggingface.co/datasets/axiong/pmc\_oa} \\
					
					PMC-VQA~\cite{Pmc-vqa} & Multi-modal & \href{https://huggingface.co/datasets/xmcmic/PMC-VQA}{https://huggingface.co/datasets/xmcmic/PMC-VQA} \\
					
					SA-Med2D-20M~\cite{SA-Med2D-20M} & Multi-modal & \href{https://openxlab.org.cn/datasets/GMAI/SA-Med2D-20M}{https://openxlab.org.cn/datasets/GMAI/SA-Med2D-20M} \\

					\bottomrule
				\end{tabular}
			}
		\end{table*}
		
		\newpage
		
		\begin{table*}
			\setlength{\abovecaptionskip}{0.pt}
			\setlength{\belowcaptionskip}{-0.em}
			\centering
			\footnotesize
			\renewcommand{\arraystretch}{1.35}
			\caption{Descriptions of datasets used in this work. We present the modalities, regions of interest, and number of triplets (image-text-annotation) in the table.}
			\begin{threeparttable}
				\resizebox{1\textwidth}{!}{
					\begin{tabular}{llll}
						\toprule[1.2pt]
						\rowcolor{lightgray}  \textbf{Dataset} &\textbf{Modality} &\textbf{Regions of interest} &\textbf{Number of triplets}\\
						\hline
						CXRMask~\cite{CXRMask} &X-Ray &Chest &1,698\\
						
						Radiography-Lung-opacity~\cite{chowdhury2020can} &X-Ray &Chest &6,012\\
						
						Radiography-Normal~\cite{chowdhury2020can} &X-Ray &Chest &30,574\\
						
						Radiography-Viral-Pneumonia~\cite{chowdhury2020can} &X-Ray &Chest &1,345\\
						
						COVID-QU-Ex~\cite{COVID-QU-Ex} &X-Ray &Chest \& COVID19 infection &20,385\\
						
						CDD-CESM~\cite{CDD-CESM} &X-Ray &Breast lesion &1,233\\
						
						SIIM~\cite{SIIM} &X-Ray &Pneumothorax &2,669\\
						
						PanNuke~\cite{pannuke} &Pathology &Connective tissue, inflammatory, neoplastic, and dead cells &32,450\\
						
						GlaS~\cite{glas} &Pathology &Gland tissue &330\\
						
						CoCaHis~\cite{cocahis} &Pathology &Colon cancer &82\\
						
						CryoNuSeg~\cite{cryonuseg} &Pathology &Nuclei &30\\
						
						DigestPath~\cite{DigestPath} &Pathology &Malignant lesion in colon tissue &13,391\\
						
						SICAPv2~\cite{SICAPv2} &Pathology &Prostate cancer &23,924\\
						
						WSSS4LUAD~\cite{WSSS4LUAD} &Pathology &Lung tumor &5,735\\
						
						CRAG~\cite{CRAG} &Pathology &Gland tissue &1,750\\
						
						MoNuSeg~\cite{MoNuSeg} &Pathology &Nuclei &51\\
						
						BreastUS~\cite{US} &Ultrasound &Breast lesion &1,294\\
						
						LiverUS~\cite{US} &Ultrasound &Liver &39\\
						
						CAMUS~\cite{CAMUS} &Ultrasound &Heart &42,463\\
						
						FH-PS-AOP~\cite{FH-PS-AOP} &Ultrasound &Transperineal &8,000\\
						
						REFUGE~\cite{refuge} &Fundus &Retinal &2,400\\
						
						DRIVE~\cite{DRIVE} &Fundus &Retinal &20\\
						
						UWater~\cite{UWaterlooSkinCancer} &Dermoscopy & Skin lesion &325\\
						
						NeoPolyp~\cite{NeoPolyp} &Endoscope &Colon polyp &2,050\\
						
						OCT-CME~\cite{OCT-CME} &OCT &Retinal edema &1,460\\
						
						BTCV~\cite{btcv} &CT &Abdomen organs &12,176 \\
						
						AMOS22~\cite{amos} &CT &Abdomen organs &138,371 \\
						
						WORD~\cite{word} &CT &Abdomen organs &58,898 \\
						
						FLARE22~\cite{FLARE} &CT &Abdomen organs &26,802 \\
						
						FLARE23~\cite{FLARE} &CT &Abdomen organs &4,760,889 \\
						
						Abdomenct1k~\cite{AbdomenCT-1K} &CT &Abdomen organs &1,006,170  \\
						
						AbdomenAtlas~\cite{atlas} &CT &Abdomen organs &3,722,697 \\
						
						AVT~\cite{AVT} &CT &Aorta &4,988 \\
						
						CHAOS~\cite{CHAOS} &CT &Liver &2,341 \\
						
						Sliver07~\cite{SLIVER07} &CT &Liver &2,750 \\
						
						IRCADb~\cite{3D-IRCADb} &CT &Liver \& Liver Tumor &3,420 \\
						
						HCCTACE~\cite{3D-IRCADb} &CT &Liver \& Liver Tumor &9,947 \\
						
						KiTS~\cite{KiTS} &CT &Kidney \& Kidney Tumor &44,557 \\
						
						TCIAPancreas~\cite{tcia-panc} &CT &Pancreas &6,882\\
						
						PANORAMA~\cite{PANORAMA} &CT &Pancreas \& Pancreas Tumor &477,437 \\
						
						QUBIQ~\cite{Qubiq} &CT &Pancreas \& Pancreas Tumor &809\\
						
						LIDC-IDRI~\cite{LIDC} &CT &Lung nodule &9,122\\
						
						COVID-CT~\cite{COVID} &CT &COVID &1,572\\
						
						RIDER~\cite{RIDER-LungCT-Seg} &CT &Lung Tumor &1,484\\
						
						AIIB23~\cite{AIIB} &CT &Fibrotic Lung disease &51,715\\
						
						MSD03 Liver~\cite{msd} &CT &Liver \& Liver Tumor &21,810\\
						MSD06 Lung~\cite{msd} &CT &Lung Tumor &1,483\\
						MSD07 Panc.~\cite{msd} &CT &Pancreas \& Pancreas Tumor &10,695\\
						MSD08 Vessel~\cite{msd} &CT &Liver Vessel Tumor &13,201\\
						MSD09 Spleen~\cite{msd} &CT &Spleen &982\\
						MSD10 Colon~\cite{msd} &CT &Colon Tumor &1,245\\
						RadGenome~\cite{ctrate,radgenome} &CT &Abdomen\&Chest &81,257\\

						
						AMOS-MRI~\cite{amos} &MRI &Abdomen &52,625\\
						
						ACDC~\cite{ACDC} &MRI &Heart &7,666\\
						
						LGG~\cite{LGG} &MRI &Brain Tumor &2,542\\
						
						
						MSD01 Brain~\cite{msd} &MRI &Brain Tumor &380,720\\
						MSD02 Heart~\cite{msd} &MRI &Heart &1,207\\
						
						MSD04 Hip.~\cite{msd} &MRI &Hippocampus  &7,770\\
						MSD05 Pros.~\cite{msd} &MRI &Prostate &1,434\\
						MedTrinity~\cite{medtrinity} &Multi-modal &Whole body &16,270,486\\
						\hline
						\textbf{Total} & & &\textbf{27,408,704}\\
						\toprule[1.2pt]
					\end{tabular}
				}
			\end{threeparttable}         
			\label{table_dataset_descriptions}
		\end{table*}
		
		\newpage
		
		\begin{table*}
			\setlength{\abovecaptionskip}{0.pt}
			\setlength{\belowcaptionskip}{-0.em}
			\centering
			\footnotesize
			\renewcommand{\arraystretch}{1.6}
			\caption{Number of images and image-text-annotation triplets across different imaging modalities. Specifically, an image may contain numerous regions of interest, and each region of interest is paired with text descriptions and localization annotations (segmentation masks or bounding boxes).}
			\begin{threeparttable}
				\resizebox{0.7\textwidth}{!}{
					\begin{tabular}{lll}
						\toprule[1.2pt]
						\rowcolor{lightgray} \textbf{Modality} &\textbf{Number of images} &\textbf{Number of triplets}\\
						\hline
						
						
						
						
						
						
						
						
						
						
						X-Ray &79,212 &122,613\\
						
						Pathology &4,029,165 &4,056,746\\
						
						CT &8,171,448 &16,193,612\\
						
						MRI &6,630,101 &6,956,286\\
						
						Ultrasound &26,066 &51,944\\
						
						Fundus &1,436 &2,636\\
						
						Dermoscopy &6,616 &6,735\\
						
						Endoscope &4,061 &5,111\\
						
						OCT &1,484 &1,484\\
						
						PET &11,561 &11,561\\
						
						\hline
						\textbf{Total} &\textbf{18,961,150} &\textbf{27,408,704}\\
						\toprule[1.2pt]
					\end{tabular}
				}
			\end{threeparttable}         
			\label{table_dataset_descriptions_modality}
		\end{table*}
		
		\clearpage
		
		\begin{table*}[!tbh]
			\renewcommand{\arraystretch}{1.6} 
			\center
			\caption{270K images with 15 types of abnormalities for grounded disease recognition experiments. 66,226 normal images as negative samples are also included in training and validation, constructing a dataset of 341,284 images. We split the datasets into $80\%$ training and $20\%$ test sets. The links to the datasets are provided in Table~\ref{table:dataset}.}
			\label{table:dataset_grounded_disease}
			\resizebox{0.9\textwidth}{!}{
				\begin{tabular}{lll}
					\toprule
					\rowcolor{lightgray} \textbf{Abnormality} & \textbf{Number} & \textbf{Datasets} \\
					\midrule
					Liver tumor	&10,338 &MSD03-Liver~\cite{msd}\\
					Pancreas tumor	&2,504 &MSD07-Pancreas~\cite{msd}\\
					Kidney tumor &12,182 &KiTS23~\cite{KiTS} \\
					
					Colon cancer	&14,718 &MSD10-Colon~\cite{msd}, CoCaHis~\cite{cocahis}, DigestPath~\cite{DigestPath}\\
					
					Lung tumor	&1,483 &MSD06-Lung~\cite{msd}\\
					
					Lung nodule	&9,122 &LIDC~\cite{LIDC} \\ 
					
					COVID19 infection	&21,957 &COVID-QU-Ex~\cite{COVID-QU-Ex}, COVID-CT~\cite{COVID}\\
					
					Fibrotic lung disease	&51,715 &AIIB~\cite{AIIB}\\
					
					Brain tumor	&119,523 &LGG~\cite{LGG}, MSD01-Brain~\cite{msd}\\
					
					Breast lesion	&1,294 &BreastUS~\cite{US}\\
					
					Colon polyp	&2,050 &NeoPolyp~\cite{NeoPolyp}\\
					
					Pneumothorax	&2,669 &Pneumonia~\cite{chowdhury2020can} \\
					
					Prostate cancer	&23,924 &SICAPv2~\cite{SICAPv2}\\
					
					Skin lesion	&119 &UWater~\cite{UWaterlooSkinCancer}\\
					
					Retinal lesion	&1,460 &OCT-CME~\cite{OCT-CME}\\
					
					No findings &66,226 &CXRMask~\cite{CXRMask}, CAMUS~\cite{CAMUS}, FH-PS-AOP~\cite{FH-PS-AOP}\\
					
					& &MSD~\cite{msd}, CHAOS~\cite{CHAOS}, KiTS23~\cite{KiTS}\\
					
					\hline
					\textbf{Total} &\textbf{341,284} &\\

					\bottomrule
				\end{tabular}
			}
		\end{table*}
		
		\newpage

		\begin{table*}[!tbh]
			\renewcommand{\arraystretch}{1.05} 
			\center
			\caption{\textbf{Internal validation: Dataset-wise segmentation dice scores on 46 datasets}. We compare our results with MedSAM~\cite{medsam} (Oracle Box as prompts), SegVol~\cite{segvol}, SAT~\cite{SAT}, and BiomedParse~\cite{biomedparse}. `/' represents that this method is not applicable to this dataset. Specifically, MedSAM and SegVol use visual prompts, which require extra inputs (bounding boxes or points) to guide the segmentation and are not applicable to datasets with dense objects. SAT and BiomedParse adopt textual prompts. SegVol and SAT are only applicable to 3D medical images. We present the confidence ranges, which reflect the results of the worst and best runs from five experimental trials. The best results are \textbf{bolded} while the second-best results are \underline{underlined}.}
			\label{table:internal_dsc}
			\resizebox{1.0\textwidth}{!}{
				\begin{tabular}{ll|ccccc}
					\toprule
					\rowcolor{lightgray} \textbf{Dataset} &\textbf{Modality} & \textbf{MedSAM} & \textbf{SegVol} & \textbf{SAT} & \textbf{BiomedParse} & \textbf{UniBiomed} \\
					\midrule
					CXRMask~\cite{CXRMask} &X-Ray &94.0(93.8,94.2) &/ &/ &\underline{96.2(96.0,96.4)} &\textbf{96.2(96.0,96.5)}\\
					COVID~\cite{chowdhury2020can} &X-Ray &91.8(91.7,92.0) &/ &/ &\underline{95.9(95.7,96.9)} &\textbf{97.1(97.0,97.2)}\\
					COVID-QU-Ex~\cite{COVID-QU-Ex} &X-Ray &87.5(87.0,88.0) &/ &/ &\underline{90.0(89.2,90.1)} &\textbf{93.6(93.4,93.8)}\\
					Lung-Opacity~\cite{chowdhury2020can} &X-Ray &90.2(90.0,90.5) &/ &/ &\underline{93.3(93.0,93.6)} &\textbf{95.9(95.8,96.0)}\\
					Pneumonia~\cite{chowdhury2020can} &X-Ray &91.5(91.2,91.7) &/ &/ &\underline{93.0(92.6,93.7)} &\textbf{96.7(96.5,96.9)}\\
					CDD-CESM~\cite{CDD-CESM} &X-Ray &/ &/ &/ &\underline{37.5(35.2,39.7)} &\textbf{40.5(40.2,40.7)}\\
					SIIM~\cite{SIIM} &X-Ray &/ &/ &/ &\underline{51.7(50.2,53.0)} &\textbf{61.0(60.7,61.3)}\\
					
					BreastUS~\cite{US} &US &76.9(76.2,77.3) &/ &/ &\underline{82.4(80.2,84.1)} &\textbf{88.8(88.2,89.4)}\\
					LiverUS~\cite{US} &US &77.4(77.0,77.8) &/ &/ &\underline{58.9(50.2,63.0)} &\textbf{82.3(81.4,83.0)}\\
					CAMUS~\cite{CAMUS} &US &78.9(78.4,80.0) &/ &/ &\underline{91.0(90.2,92.5)} &\textbf{94.4(93.8,95.0)}\\
					FH-PS-AOP~\cite{FH-PS-AOP} &US &75.3(73.8,77.0) &/ &/ &\underline{81.0(80.0,81.9)} &\textbf{83.9(83.7,84.1)}\\
					
					PanNuke~\cite{pannuke} &Path. &/ &/ &/ &\underline{58.9(57.8,60.0)} &\textbf{63.5(63.2,64.3)}\\
					GlaS~\cite{glas} &Path. &/ &/ &/ &\underline{87.9(86.8,89.0)} &\textbf{96.1(95.9,96.3)}\\
					CoCaHis~\cite{cocahis} &Path. &/ &/ &/ &\underline{51.3(50.2,52.0)} &\textbf{70.5(69.8,71.0)}\\
					CryoNuSeg~\cite{cryonuseg} &Path. &/ &/ &/ &\underline{73.5(73.1,74.3)} &\textbf{82.1(81.8,82.4)}\\
					DigestPath~\cite{DigestPath} &Path. &/ &/ &/ &\underline{53.5(53.0,54.1)} &\textbf{69.9(69.8,70.0)}\\
					SICAPv2~\cite{SICAPv2} &Path. &/ &/ &/ &\underline{60.2(58.9,61.4)} &\textbf{68.4(68.0,68.9)}\\
					WSSS4LUAD~\cite{WSSS4LUAD} &Path. &/ &/ &/ &\underline{68.8(67.9,70.0)} &\textbf{76.3(76.1,76.6)}\\
					
					REFUGE~\cite{refuge} &Fundus &54.5(50.8,58.0) &/ &/ &\underline{89.6(88.7,90.2)} &\textbf{89.9(89.3,90.2)}\\
					DRIVE~\cite{DRIVE} &Fundus &/ &/ &/ &\underline{32.8(23.8,41.1)} &\textbf{71.7(70.8,72.6)}\\
					
					UWater~\cite{UWaterlooSkinCancer} &Derm. &86.8(85.9,87.7) &/ &/ &\underline{92.1(91.8,92.5)} &\textbf{93.6(93.3,94.0)}\\
					NeoPolyp~\cite{NeoPolyp} &Endo. &79.5(75.8,81.0) &/ &/ &\underline{91.7(89.7,92.5)} &\textbf{93.6(93.2,94.1)}\\
					OCT-CME~\cite{OCT-CME} &OCT &71.6(70.5,72.4) &/ &/ &\underline{85.6(83.8,87.0)} &\textbf{87.4(86.8,88.0)}\\
					
					BTCV~\cite{btcv} & CT &/ &77.6(76.5,78.4) &79.6(79.5,79.8) &70.1(69.5,70.4) &\textbf{83.9(83.5,84.4)}\\
					WORD~\cite{word} & CT &/ &83.2(82.4,83.8) &\textbf{86.5(85.6,87.4)} &56.2(51.6,60.4) &\underline{86.2(85.9,86.8)}\\
					FLARE22~\cite{FLARE}  &CT &/ &87.2(86.5,88.9) &\textbf{88.8(88.5,89.1)} &71.3(70.6,72.2) &\underline{87.6(87.0,88.4)}\\
					KiTS23~\cite{KiTS} & CT &67.9(66.5,69.3) &66.9(66.5,67.4) &\underline{67.9(66.5,69.4)} &66.3(66.0,66.8) &\textbf{77.8(77.6,78.2)}\\
					MSD03-Liver~\cite{msd} & CT &/ &78.2(77.8,78.6) &68.1(67.5,68.4) &\underline{83.8(83.3,84.2)} &\textbf{84.7(83.5,85.4)}\\
					MSD06-Lung~\cite{msd} & CT &31.3(24.5,42.0) &\underline{55.6(55.0,56.4)} &51.0(50.5,51.5) &55.4(54.5,56.2) &\textbf{56.0(55.8,56.4)}\\
					MSD07-Panc.~\cite{msd} & CT &/ &40.8(40.5,41.2) &\underline{54.2(53.5,55.4)} &31.1(26.6,34.4) &\textbf{54.8(53.9,55.4)}\\
					MSD08-Hepatic~\cite{msd} & CT &/ &/ &53.5(50.5,57.0) &\underline{57.2(56.0,58.1)} &\textbf{59.2(58.5,59.9)}\\
					MSD09-Spleen~\cite{msd} & CT &91.1(90.4,91.6) &91.7(91.4,92.3) &\textbf{93.5(92.6,94.3)} &87.9(86.4,89.0) &\underline{92.2(92.0,92.5)}\\
					MSD10-Colon~\cite{msd} & CT &/ &37.6(29.4,43.3) &35.3(34.4,37.0) &\underline{39.4(38.4,40.3)} &\textbf{43.6(43.4,43.8)}\\
					
					LIDC-IDRI~\cite{LIDC} &CT &/ &24.6(21.4,28.7) &28.0(27.0,29.0) &\underline{59.9(59.4,60.3)} &\textbf{73.2(73.0,73.7)}\\
					AIIB~\cite{AIIB} &CT &/ &/ &\underline{34.0(28.0,37.6)} &31.2(30.4,32.3) &\textbf{64.6(64.1,65.3)}\\
					COVID-CT~\cite{COVID} &CT &58.2(57.4,59.3) &\underline{75.8(75.4,76.3)} &71.5(70.4,72.6) &73.7(73.4,74.0) &\textbf{88.8(88.3,89.3)}\\
					AMOS-CT~\cite{amos} &CT &81.2(80.4,82.0) &80.9(80.4,81.3) &84.9(84.4,85.1) &\underline{88.2(87.9,88.6)} &\textbf{91.4(91.0,91.9)}\\
					AVT~\cite{AVT} &CT &/ &\underline{49.2(48.4,50.0)} &45.5(44.4,46.3) &22.0(17.4,26.3) &\textbf{52.4(51.6,53.3)}\\
					RadGenome~\cite{radgenome} &CT &/ &26.6(22.4,27.3) &15.2(14.4,17.6) &\underline{34.2(31.4,38.3)} &\textbf{53.2(51.4,54.3)}\\
					
					LGG~\cite{LGG}  &MRI &/ &/ &75.6(74.4,76.0) &\underline{89.5(89.0,90.1)} &\textbf{93.2(93.0,93.5)}\\
					MSD01-Brain~\cite{msd} & MRI &/ &/ &/ &\underline{70.5(70.2,70.9)} &\textbf{70.6(70.3,70.9)}\\
					MSD02-Heart~\cite{msd} & MRI &60.8(59.7,61.0) &/ &\underline{90.2(89.7,90.9)} &86.1(85.7,86.8) &\textbf{90.5(89.7,91.1)}\\
					MSD04-Hip.~\cite{msd} & MRI &/ &/ &\underline{82.0(81.8,82.3)} &80.4(79.7,80.9) &\textbf{82.1(81.9,82.3)}\\
					MSD05-Pros.~\cite{msd} & MRI &\underline{69.1(68.8,69.3)} &/ &68.3(67.7,69.0) &36.8(33.0,38.4) &\textbf{70.0(68.8,72.3)}\\
					ACDC~\cite{ACDC}  &MRI &/ &/ &66.7(65.9,67.3) &\underline{87.1(86.7,87.4)} &\textbf{87.8(86.7,88.2)}\\
					AMOS-MRI~\cite{amos} &MRI &79.1(78.6,80.0) &/ &78.8(78.2,79.3) &\underline{83.2(81.7,85.1)} &\textbf{84.2(82.8,86.8)}\\
					
					\bottomrule
				\end{tabular}
			}
		\end{table*}
		
		\clearpage

		\begin{table*}[!tbh]
			\renewcommand{\arraystretch}{1.2} 
			\center
			\caption{\textbf{External validation: Dataset-wise segmentation dice scores on 14 datasets}. We compare our results with MedSAM~\cite{medsam} (Oracle Box as prompts), SAT~\cite{SAT}, and BiomedParse~\cite{biomedparse}. `/' represents that this method is not applicable to this dataset. Specifically, MedSAM uses visual prompts, while SAT, BiomedParse, and UniBiomed adopt textual prompts. Some datasets are already included in the training datasets of SAT, where we annotate with `/' symbols.
				We present the confidence ranges, which reflect the results of the worst and best runs from five experimental trials. The best results are \textbf{bolded} while the second-best results are \underline{underlined}.}
			\label{table:External_dsc}
			\resizebox{1.0\textwidth}{!}{
				\begin{tabular}{ll|cccc}
					\toprule
					\rowcolor{lightgray} \textbf{Dataset} &\textbf{Modality} & \textbf{MedSAM} & \textbf{SAT} & \textbf{BiomedParse} & \textbf{UniBiomed} \\
					\midrule
					CRAG~\cite{CRAG} &Path. &/ &/ &\underline{62.5(61.3,63.4)} &\textbf{84.0(83.7,84.5)} \\
					MoNuSeg~\cite{MoNuSeg} &Path. &/ &/ &\underline{60.2(59.5,60.8)} &\textbf{79.7(79.5,80.0)} \\
					CHAOS~\cite{CHAOS} & CT &\underline{76.5(76.0,77.0)} &/ &70.1(69.7,70.4) &\textbf{94.6(94.0,95.2)} \\
					IRCADb~\cite{3D-IRCADb}  &CT &17.1(10.3,23.2) &\underline{56.7(55.3,57.5)} &47.2(46.3,48.4) &\textbf{65.0(64.6,65.4)} \\
					SLIVER07~\cite{SLIVER07} &CT &68.9(68.0,69.6) &/ &\underline{79.8(78.3,81.4)} &\textbf{88.6(88.2,90.1)} \\
					HCCTACE~\cite{HCCTACE} &CT &16.6(11.0,20.5) &\underline{64.4(62.9,65.0)} &55.3(54.8,55.9) &\textbf{65.2(65.0,65.4)} \\
					TCIAPancreas~\cite{tcia-panc} &CT &/ &/ &\underline{61.9(61.0,62.8)} &\textbf{72.5(72.3,72.7)} \\
					QUBIQ~\cite{Qubiq} &CT &/ &\underline{40.8(39.9,41.8)} &33.3(32.3,34.0) &\textbf{46.6(46.3,46.9)} \\
					Rider~\cite{RIDER-LungCT-Seg} &CT &/ &\underline{37.6(37.0,38.2)} &31.8(30.6,33.0) &\textbf{45.0(44.8,45.2)} \\
					PANORAMA~\cite{PANORAMA} &CT &/ &\underline{30.5(30.0,31.3)} &25.1(22.7,28.4) &\textbf{36.2(35.6,37.0}) \\
					AbdomenCT1K~\cite{AbdomenCT-1K} &CT &\underline{82.8(81.3,83.4)} &/ &66.9(66.0,68.2) &\textbf{84.6(84.0,85.4)} \\
					FLARE23~\cite{FLARE}  &CT &/ &/ &\underline{38.9(37.3,40.4)} &\textbf{47.2(46.6,47.9)} \\
					AbdomenAtlas~\cite{atlas} &CT &/ &\underline{74.1(72.8,75.4)} &63.4(62.3,65.4) &\textbf{77.9(77.3,78.3)} \\
					BraTS21~\cite{brats} &MRI &/ &/ &\underline{73.8(73.1,74.4)} &\textbf{78.5(78.0,79.0)} \\
					
					\bottomrule
				\end{tabular}
			}
		\end{table*}

		\clearpage
		
		\begin{table*}[!tbh]
			\renewcommand{\arraystretch}{1.2} 
			\center
			\caption{\textbf{The dice score results of grounded disease recognition across 15 types of abnormalities}. We reproduce GLaMM~\cite{glamm} and LISA~\cite{lisa} on our datasets for comparisons. We present the confidence ranges, which reflect the results of the worst and best runs from five experimental trials. The best results are \textbf{bolded}.}
			\label{table:Grounded_disease}
			\resizebox{0.7\textwidth}{!}{
				\begin{tabular}{l|ccc}
					\toprule
					\rowcolor{lightgray} \textbf{Abnormality} & \textbf{GLaMM}  & \textbf{LISA} & \textbf{UniBiomed} \\
					\midrule
					Liver tumor & 36.3(35.0,36.8) & 37.9(36.6,38.5) & \textbf{49.2(48.4,50.2)} \\
					Pancreas tumor & 20.1(19.3,21.3) & 18.6(18.2,19.2) & \textbf{34.8(33.2,35.3)} \\
					Kidney tumor & 26.8(25.4,27.4) & 30.9(29.7,32.3) & \textbf{40.1(38.7,41.2)} \\
					Colon cancer & 58.3(57.0,60.0) & 61.2(59.8,62.8) & \textbf{67.8(66.4,68.8)} \\
					Lung tumor & 48.8(48.4,50.1) & 51.0(49.5,52.6) & \textbf{53.8(52.3,55.9)} \\
					Lung nodule & 47.2(46.5,48.1) & 50.7(49.6,51.7) & \textbf{52.0(51.7,53.2)} \\
					COVID19 infection & 63.1(62.3,63.5) & 68.5(68.4,70.0) & \textbf{72.7(72.0,73.4)} \\
					Fibrotic lung disease & 57.8(56.8,59.9) & 62.8(61.4,63.7) & \textbf{64.8(64.4,66.7)} \\
					Brain tumor & 76.5(75.5,77.3) & 78.4(77.4,79.4) & \textbf{82.3(82.5,83.7)} \\
					Breast lesion & 81.3(81.5,82.4) & 85.8(85.5,86.5) & \textbf{91.3(90.6,92.1)} \\
					Colon polyp & 84.3(83.1,85.1) & 86.7(86.8,88.0) & \textbf{88.0(86.9,88.8)} \\
					Pneumothorax & 47.9(46.8,49.2) & 45.2(42.8,46.8) & \textbf{51.4(50.0,52.0)} \\
					Prostate cancer & 66.9(66.8,67.7) & 72.1(71.4,72.9) & \textbf{76.4(74.7,77.9)} \\
					Skin lesion & 88.6(88.0,89.8) & 90.8(89.8,92.2) & \textbf{93.9(93.2,94.2)} \\
					Retinal edema & 79.1(78.4,79.6) & 80.3(79.1,80.6) & \textbf{83.4(83.0,85.0)} \\
					\bottomrule
				\end{tabular}
			}
		\end{table*}
		
		\begin{table*}[!tbh]
			\renewcommand{\arraystretch}{1.2} 
			\center
			\caption{\textbf{The grounded report generation results on RadGenome dataset~\cite{radgenome}}. We reproduce GLaMM~\cite{glamm} and LISA~\cite{lisa} on our datasets for comparisons. We report the results of dice scores, BLEU, METEOR, and ROUGE\_L. 
				We present the confidence ranges, which reflect the results of the worst and best runs from five experimental trials. The best results are \textbf{bolded}.}
			\label{table:RADGENOME}
			\resizebox{0.65\textwidth}{!}{
				\begin{tabular}{l|ccc}
					\toprule
					\rowcolor{lightgray} \textbf{Metric} & \textbf{GLaMM}  & \textbf{LISA} & \textbf{UniBiomed} \\
					\midrule
					Dice score & 48.6(46.4,48.9) & 51.3(49.3,51.5) & \textbf{53.8(52.7,55.0)} \\
					BLEU\_1 & 41.7(41.4,42.6) & 40.2(39.8,41.5) & \textbf{45.7(44.2,46.7)} \\
					BLEU\_2 & 36.5(35.2,37.2) & 32.3(31.9,32.9) & \textbf{38.8(38.5,39.4)} \\
					BLEU\_3 & 31.2(29.5,32.6) & 26.9(26.2,28.5) & \textbf{33.7(32.7,34.7)} \\
					BLEU\_4 & 28.1(27.7,28.8) & 24.1(23.2,25.6) & \textbf{30.0(28.8,30.6)} \\
					METEOR & 21.0(20.4,21.7) & 21.0(20.5,22.1) & \textbf{24.8(24.2,25.5)} \\
					ROUGE\_L & 49.0(48.8,50.5) & 45.3(43.8,46.2) & \textbf{51.6(50.1,52.3)} \\
					\bottomrule
				\end{tabular}
			}
		\end{table*}
		
		\clearpage
		
		\begin{table*}[!tbh]
			\renewcommand{\arraystretch}{1.2} 
			\center
			\caption{\textbf{The average accuracy results of ROI classification across 10 modalities}. The datasets for each modality are presented in Extended Data Table~\ref{table:dataset}. We compare UniBiomed with MedRegA~\cite{MedRegA} and MedPLIB~\cite{MedPLIB}. We present the confidence ranges, which reflect the results of the worst and best runs from five experimental trials. The best results are \textbf{bolded}.}
			\label{table:ROI}
			\resizebox{0.7\textwidth}{!}{
				\begin{tabular}{l|ccc}
					\toprule
					\rowcolor{lightgray} \textbf{Modality} & \textbf{MedRegA}  & \textbf{MedPLIB} & \textbf{UniBiomed} \\
					\midrule
					CT & 72.3(70.9,73.4) & 75.0(74.2,74.9) & \textbf{85.5(84.6,85.9)} \\
					MRI & 66.1(64.9,66.5) & 69.8(69.1,70.8) & \textbf{78.9(78.2,79.5)} \\
					X-ray & 79.1(78.6,79.6) & 84.8(85.6,85.7) & \textbf{93.2(92.8,93.4)} \\
					Pathology & 68.8(67.9,68.9) & 73.2(72.4,73.0) & \textbf{81.8(81.8,82.7)} \\
					Ultrasound & 93.3(92.8,93.0) & 93.8(92.3,94.4) & \textbf{98.4(96.9,99.3)} \\
					Fundus & 90.6(90.1,90.7) & 92.4(92.3,92.9) & \textbf{98.5(97.3,99.5)} \\
					Dermoscopy & 52.1(52.1,52.4) & 51.6(50.4,51.6) & \textbf{90.8(90.3,91.9)} \\
					Endoscope & 91.0(89.9,91.6) & 93.0(92.5,93.6) & \textbf{95.5(94.9,96.2)} \\
					OCT & 92.4(92.5,93.0) & 98.0(97.8,98.1) & \textbf{98.0(97.3,98.1)} \\
					PET & 93.0(90.1,90.8) & 86.8(85.9,88.4) & \textbf{92.5(91.8,93.2)} \\
					\bottomrule
				\end{tabular}
			}
		\end{table*}
		
		\begin{table*}[!tbh]
			\renewcommand{\arraystretch}{1.2} 
			\center
			\caption{\textbf{The report generation results on MedTrinity dataset~\cite{medtrinity}}. We compare UniBiomed with InternVL2.5~\cite{internvl}, LLaVA-Med~\cite{Llava-med}, MedRegA~\cite{MedRegA} and MedPLIB~\cite{MedPLIB}. We report the results of BLEU, METEOR, and ROUGE\_L. 
				We present the confidence ranges, which reflect the results of the worst and best runs from five experimental trials. The best results are \textbf{bolded}.}
			\label{table:medtrinity}
			\resizebox{0.9\textwidth}{!}{
				\begin{tabular}{l|ccccc}
					\toprule
					\rowcolor{lightgray} \textbf{Metric} & \textbf{InternVL}  & \textbf{LLaVA-Med} & \textbf{MedRegA}  & \textbf{MedPLIB} & \textbf{UniBiomed} \\
					\midrule
					BLEU\_1 & 24.9(23.4,26.4) & 27.0(25.0,28.0) & 41.7(40.8,42.6) & 43.6(41.4,45.8) & \textbf{52.4(50.5,53.1)} \\
					BLEU\_2 & 14.0(12.5,15.4) & 16.0(15.2,17.5) & 30.1(28.3,31.6) & 31.0(28.7,31.7) & \textbf{46.0(44.1,47.0)} \\
					BLEU\_3 & 7.7(6.1,9.3) & 17.9(16.2,20.2) & 26.2(24.9,28.2) & 28.0(26.2,29.9) & \textbf{37.9(36.8,39.3)} \\
					BLEU\_4 & 3.9(2.3,5.2) & 12.0(10.1,13.8) & 21.5(19.7,22.6) & 22.0(22.3,23.5) & \textbf{32.1(30.0,33.6)} \\
					METEOR & 17.6(16.0,19.3) & 23.2(22.1,24.8) & 23.7(22.3,25.0) & 28.1(27.4,30.3) & \textbf{30.4(30.5,32.5)} \\
					ROUGE\_L & 18.6(17.2,19.8) & 18.5(16.9,21.0) & 27.4(25.7,29.0) & 32.1(30.1,34.5) & \textbf{47.9(46.8,49.2)} \\
					\bottomrule
				\end{tabular}
			}
		\end{table*}
		
		\clearpage

		\begin{table*}[!tbh]
			\renewcommand{\arraystretch}{0.7} 
			\centering
			\caption{\rebut{The detailed statistics for 46 internal segmentation datasets. The mean dice scores, confidence intervals, effect size, and p-values compared with the best-competing method~\cite{biomedparse} are reported. We report the confidence interval via five experimental trials. For comparisons, we evaluate the paired t-test for each dataset, and we report the mean values of improvements and the p-values.}}
			\label{table:internal_dsc_stats}
			\resizebox{0.9\textwidth}{!}{
				\begin{tabular}{ll|cc|ccc}
					\toprule
					\rowcolor{lightgray} \textbf{Dataset} & \textbf{Modality} & \textbf{BiomedParse} & \textbf{UniBiomed}  & \textbf{Effect size} & \textbf{Mean improve.} & \textbf{$P$-values}\\
					\midrule
					CXRMask~\cite{CXRMask} & X-Ray & 96.2(96.0,96.4) & 96.2(96.0,96.5) &66 & 0.0 & $4.10\times10^{-1}$\\
					COVID~\cite{chowdhury2020can} & X-Ray & 95.9(95.7,96.9) & 97.1(97.0,97.2) &2,172 & 1.2 & $1.09\times10^{-17}$\\
					COVID-QU-Ex~\cite{COVID-QU-Ex} & X-Ray & 90.0(89.2,90.1) & 93.6(93.4,93.8) &4,079 & 3.6 & $3.02\times10^{-5}$\\
					Lung-Opacity~\cite{chowdhury2020can} & X-Ray & 93.3(93.0,93.6) & 95.9(95.8,96.0) &1,203 & 2.6 &$2.48\times10^{-2}$ \\
					Pneumonia~\cite{chowdhury2020can} & X-Ray & 93.0(92.6,93.7) & 96.7(96.5,96.9) &269 & 3.7 &$7.46\times10^{-2}$ \\
					CDD-CESM~\cite{CDD-CESM} & X-Ray & 37.5(35.2,39.7) & 40.5(40.2,40.7) &217 & 3.0 &$9.76\times10^{-2}$ \\
					SIIM~\cite{SIIM} & X-Ray & 51.7(50.2,53.0) & 61.0(60.7,61.3) &290 & 9.3 &$1.60\times10^{-34}$ \\
					BreastUS~\cite{US} & US & 82.4(80.2,84.1) & 88.8(88.2,89.4) &256 & 6.4 &$8.69\times10^{-11}$ \\
					LiverUS~\cite{US} & US & 58.9(50.2,63.0) & 82.3(81.4,83.0) &10 & 23.4 &$1.01\times10^{-3}$ \\
					CAMUS~\cite{CAMUS} & US & 91.0(90.2,92.5) & 94.4(93.8,95.0) &8,328 & 3.4 &$2.30\times10^{-26}$ \\
					FH-PS-AOP~\cite{FH-PS-AOP} & US & 81.0(80.0,81.9) & 83.9(83.7,84.1) &1,600 & 2.9 &$4.70\times10^{-15}$ \\
					PanNuke~\cite{pannuke} & Path. & 58.9(57.8,60.0) & 63.5(63.2,64.3) &6,414 & 4.6 &$2.54\times10^{-5}$ \\
					GlaS~\cite{glas} & Path. & 87.9(86.8,89.0) & 96.1(95.9,96.3) &122 & 8.2 &$9.15\times10^{-6}$ \\
					CoCaHis~\cite{cocahis} & Path. & 51.3(50.2,52.0) & 70.5(69.8,71.0) &24 & 19.2 &$1.80\times10^{-17}$ \\
					CryoNuSeg~\cite{cryonuseg} & Path. & 73.5(73.1,74.3) & 82.1(81.8,82.4) &10 & 8.6 &$4.17\times10^{-11}$ \\
					DigestPath~\cite{DigestPath} & Path. & 53.5(53.0,54.1) & 69.9(69.8,70.0) &2,666 & 16.4 &$3.56\times10^{-12}$ \\
					SICAPv2~\cite{SICAPv2} & Path. & 60.2(58.9,61.4) & 68.4(68.0,68.9) &4,784 & 8.2 &$5.18\times10^{-7}$ \\
					WSSS4LUAD~\cite{WSSS4LUAD} & Path. & 68.8(67.9,70.0) & 76.3(76.1,76.6) &1,145 & 7.5 &$2.21\times10^{-8}$ \\
					REFUGE~\cite{refuge} & Fundus & 89.6(88.7,90.2) & 89.9(89.3,90.2) &800 & 0.3 &$8.40\times10^{-3}$ \\
					DRIVE~\cite{DRIVE} & Fundus & 32.8(23.8,41.1) & 71.7(70.8,72.6) &10 & 38.9 &$1.48\times10^{-7}$ \\
					UWater~\cite{UWaterlooSkinCancer} & Derm. & 92.1(91.8,92.5) & 93.6(93.3,94.0) &65 & 1.5 &$6.14\times10^{-2}$ \\
					NeoPolyp~\cite{NeoPolyp} & Endo. & 91.7(89.7,92.5) & 93.6(93.2,94.1) &410 & 1.9 &$3.67\times10^{-5}$ \\
					OCT-CME~\cite{OCT-CME} & OCT & 85.6(83.8,87.0) & 87.4(86.8,88.0) &283 & 1.8 &$1.70\times10^{-3}$ \\
					BTCV~\cite{btcv} & CT & 70.1(69.5,70.4) & 83.9(83.5,84.4) &2,527 & 13.8 &$1.11\times10^{-27}$ \\
					WORD~\cite{word} & CT & 56.2(51.6,60.4) & 86.2(85.9,86.8) &11,725 & 30.0 &$4.70\times10^{-22}$ \\
					FLARE22~\cite{FLARE} & CT & 71.3(70.6,72.2) & 87.6(87.0,88.4) &5,515 & 16.3 &$9.01\times10^{-14}$ \\
					KiTS23~\cite{KiTS} & CT & 66.3(66.0,66.8) & 77.8(77.6,78.2) &9,073 & 11.5 &$7.76\times10^{-12}$ \\
					MSD03-Liver~\cite{msd} & CT & 83.8(83.3,84.2) & 84.7(83.5,85.4) &5,621 & 0.9 &$8.99\times10^{-3}$ \\
					MSD06-Lung~\cite{msd} & CT & 55.4(54.5,56.2) & 56.0(55.8,56.4) &242 & 0.6 &$4.54\times10^{-1}$ \\
					MSD07-Panc.~\cite{msd} & CT & 31.1(26.6,34.4) & 54.8(53.9,55.4) &2,210 & 23.7 &$3.35\times10^{-21}$ \\
					MSD08-Hepatic~\cite{msd} & CT & 57.2(56.0,58.1) & 59.2(58.5,59.9) &3,122 & 2.0 &$4.08\times10^{-6}$ \\
					MSD09-Spleen~\cite{msd} & CT & 87.9(86.4,89.0) & 92.2(92.0,92.5) &198 & 4.3 &$7.70\times10^{-2}$ \\
					MSD10-Colon~\cite{msd} & CT & 39.4(38.4,40.3) & 43.6(43.4,43.8) &245 & 4.2 &$4.89\times10^{-32}$ \\
					LIDC-IDRI~\cite{LIDC} & CT & 59.9(59.4,60.3) & 73.2(73.0,73.7) &1,733 & 13.3 &$3.01\times10^{-4}$ \\
					AIIB~\cite{AIIB} & CT & 31.2(30.4,32.3) & 64.6(64.1,65.3) &10,704 & 33.4 &$1.87\times10^{-32}$ \\
					COVID-CT~\cite{COVID} & CT & 73.7(73.4,74.0) & 88.8(88.3,89.3) &385 & 15.1 &$3.34\times10^{-33}$ \\
					AMOS-CT~\cite{amos} & CT & 88.2(87.9,88.6) & 91.4(91.0,91.9) &29,665 & 3.2 &$5.22\times10^{-3}$ \\
					AVT~\cite{AVT} & CT & 22.0(17.4,26.3) & 52.4(51.6,53.3) &263 & 30.4 & $8.68\times10^{-15}$\\
					RadGenome~\cite{radgenome} & CT & 34.2(31.4,38.3) & 53.2(51.4,54.3) &5,100 & 19.0 &$6.36\times10^{-10}$ \\
					LGG~\cite{LGG} & MRI & 89.5(89.0,90.1) & 93.2(93.0,93.5) &514 & 3.7 &$9.63\times10^{-18}$ \\
					MSD01-Brain~\cite{msd} & MRI & 70.5(70.2,70.9) & 70.6(70.3,70.9) &76,432 & 0.1 &$2.61\times10^{-11}$ \\
					MSD02-Heart~\cite{msd} & MRI & 86.1(85.7,86.8) & 90.5(89.7,91.1) &248 & 4.4 &$2.86\times10^{-5}$ \\
					MSD04-Hip.~\cite{msd} & MRI & 80.4(79.7,80.9) & 82.1(81.9,82.3) &1,562 & 1.7 & $7.50\times10^{-3}$\\
					MSD05-Pros.~\cite{msd} & MRI & 36.8(33.0,38.4) & 70.0(68.8,72.3) &308 & 33.2 &$9.86\times10^{-27}$ \\
					ACDC~\cite{ACDC} & MRI & 87.1(86.7,87.4) & 87.8(86.7,88.2) &2,732 & 0.7 &$2.98\times10^{-13}$ \\
					AMOS-MRI~\cite{amos} & MRI & 83.2(81.7,85.1) & 84.2(82.8,86.8) &8,150 & 1.0 &$1.39\times10^{-2}$ \\
					\bottomrule
				\end{tabular}
			}
		\end{table*}
		
		\begin{table*}[!tbh]
			\renewcommand{\arraystretch}{1} 
			\centering
			\caption{\rebut{The detailed statistics for 14 external segmentation datasets. The mean dice scores, confidence intervals, effect size, and p-values compared with the best-competing method~\cite{biomedparse} are reported. We report the confidence interval via five experimental trials. For comparisons, we evaluate the paired t-test for each dataset, and we report the mean values of improvements and the p-values.}}
			\label{table:External_dsc_stats}
			\resizebox{0.7\textwidth}{!}{
				\begin{tabular}{ll|cc|ccc}
					\toprule
					\rowcolor{lightgray} \textbf{Dataset} &\textbf{Modality} & \textbf{BiomedParse} & \textbf{UniBiomed} & \textbf{Effect size} & \textbf{Mean improve.} & \textbf{$P$-values}\\
					\midrule
					CRAG~\cite{CRAG} &Path. &62.5(61.3,63.4) &84.0(83.7,84.5) &321 &21.5 &$5.79\times10^{-58}$\\
					MoNuSeg~\cite{MoNuSeg} &Path. &60.2(59.5,60.8) &79.7(79.5,80.0) &14 &19.5 &$1.10\times10^{-5}$\\
					CHAOS~\cite{CHAOS} & CT &70.1(69.7,70.4) &94.6(94.0,95.2) &529 &24.5 &$1.23\times10^{-3}$\\
					IRCADb~\cite{3D-IRCADb}  &CT &47.2(46.3,48.4) &65.0(64.6,65.4) &574 &17.8 &$1.73\times10^{-9}$\\
					SLIVER07~\cite{SLIVER07} &CT &79.8(78.3,81.4) &88.6(88.2,90.1) &644 &8.8 &$5.43\times10^{-40}$\\
					HCCTACE~\cite{HCCTACE} &CT &55.3(54.8,55.9) &65.2(65.0,65.4) &1,508 &9.9 &$3.37\times10^{-6}$\\
					TCIAPancreas~\cite{tcia-panc} &CT &61.9(61.0,62.8) &72.5(72.3,72.7) & 1,333 &10.6 &$2.93\times10^{-34}$\\
					QUBIQ~\cite{Qubiq} &CT &33.3(32.3,34.0) &46.6(46.3,46.9) &179 &13.3 &$6.50\times10^{-11}$\\
					Rider~\cite{RIDER-LungCT-Seg} &CT &31.8(30.6,33.0) &45.0(44.8,45.2) &332 &13.2 &$7.25\times10^{-15}$\\
					PANORAMA~\cite{PANORAMA} &CT &25.1(22.7,28.4) &36.2(35.6,37.0) &64,296 &11.1 &$7.20\times10^{-7}$\\
					AbdomenCT1K~\cite{AbdomenCT-1K} &CT &66.9(66.0,68.2) &84.6(84.0,85.4) &179,372 &17.7 &$8.08\times10^{-5}$\\
					FLARE23~\cite{FLARE}  &CT &38.9(37.3,40.4) &47.2(46.6,47.9) &976,933 &8.3 &$4.09\times10^{-3}$\\
					AbdomenAtlas~\cite{atlas} &CT &63.4(62.3,65.4) &77.9(77.3,78.3) &737,368 &14.5 &$1.78\times10^{-3}$\\
					BraTS21~\cite{brats} &MRI &73.8(73.1,74.4) &78.5(78.0,79.0) &1,215 &4.7 &$2.72\times10^{-6}$\\
					\bottomrule
				\end{tabular}
			}
		\end{table*}
		
		\begin{table*}[!tbh]
			\renewcommand{\arraystretch}{1.2} 
			\centering
			\caption{\rebut{The detailed statistics for the grounded disease recognition task. The mean dice scores, confidence intervals, effect size, and p-values compared with the best-competing method~\cite{lisa} are reported. We report the confidence interval via five experimental trials. For comparisons, we evaluate the paired t-test for each dataset, and we report the mean values of improvements and the p-values.}}
			\label{table:Grounded_disease_stats}
			\resizebox{0.7\textwidth}{!}{
				\begin{tabular}{l|cc|ccc}
					\toprule
					\rowcolor{lightgray} \textbf{Abnormality} & \textbf{LISA} & \textbf{UniBiomed} & \textbf{Effect size} & \textbf{Mean improve.} & \textbf{$P$-values}\\
					\midrule
					Liver tumor & 37.9(36.6,38.5) &49.2(48.4,50.2) &10,338 & 11.3 &$3.51\times10^{-6}$\\
					Pancreas tumor & 18.6(18.2,19.2) &34.8(33.2,35.3) &2,504 & 16.2 &$1.08\times10^{-3}$\\
					Kidney tumor & 30.9(29.7,32.3) & 40.1(38.7,41.2) &12,182 & 9.2 &$9.80\times10^{-3}$\\
					Colon cancer & 61.2(59.8,62.8) & 67.8(66.4,68.8) &14,718 & 6.6 &$1.56\times10^{-8}$\\
					Lung tumor & 51.0(49.5,52.6) & 53.8(52.3,55.9) &1,483 & 2.8 &$7.71\times10^{-3}$\\
					Lung nodule & 50.7(49.6,51.7) & 52.0(51.7,53.2) &9,122 & 1.3 &$1.60\times10^{-2}$\\
					COVID19 infection & 68.5(68.4,70.0) &72.7(72.0,73.4) &21,957 & 4.2 &$2.58\times10^{-3}$\\
					Fibrotic lung disease & 62.8(61.4,63.7) & 64.8(64.4,66.7) &51,715 & 2.0 &$7.83\times10^{-6}$\\
					Brain tumor & 78.4(77.4,79.4) & 82.3(82.5,83.7) &119,523 & 3.9 &$4.15\times10^{-3}$\\
					Breast lesion & 85.8(85.5,86.5) & 91.3(90.6,92.1) &1,294 & 5.5 &$1.46\times10^{-3}$\\
					Colon polyp & 86.7(86.8,88.0) & 88s.0(86.9,88.8) &2,050 & 1.3 &$9.77\times10^{-2}$\\
					Pneumothorax & 45.2(42.8,46.8) & 51.4(50.0,52.0) &2,669 & 6.2 &$1.25\times10^{-4}$\\
					Prostate cancer & 72.1(71.4,72.9) & 76.4(74.7,77.9) &23,924 & 4.3 &$2.09\times10^{-4}$\\
					Skin lesion & 90.8(89.8,92.2) & 93.9(93.2,94.2) &119 & 3.1 &$6.18\times10^{-3}$\\
					Retinal edema & 80.3(79.1,80.6) & 83.4(83.0,85.0) &1,460 & 3.1 &$2.480\times10^{-4}$\\
					\bottomrule
				\end{tabular}
			}
		\end{table*}
		
		\begin{table*}[!tbh]
			\renewcommand{\arraystretch}{1.2} 
			\centering
			\caption{\rebut{The detailed statistics for the ROI classification task. The mean accuracy, confidence intervals, effect size, and p-values compared with the best-competing method~\cite{MedPLIB} are reported. We report the confidence interval via five experimental trials. For comparisons, we evaluate the paired t-test for each dataset, and we report the mean values of improvements and the p-values.}}
			\label{table:ROI_stats}
			\resizebox{0.7\textwidth}{!}{
				\begin{tabular}{l|cc|ccc}
					\toprule
					\rowcolor{lightgray} \textbf{Modality} & \textbf{MedPLIB} & \textbf{UniBiomed} & \textbf{Effect size} & \textbf{Mean improve.} & \textbf{$P$-values}\\
					\midrule
					CT & 75.0(74.2,74.9) & 85.5(84.6,85.9) &2,048,264  & 10.5 &$1.06\times10^{-3}$\\
					MRI & 69.8(69.1,70.8) & 78.9(78.2,79.5) &93,068 & 9.1 &$9.11\times10^{-2}$\\
					X-ray & 84.8(85.6,85.7) & 93.2(92.8,93.4) &14,413 & 8.4 &$2.72\times10^{-3}$\\
					Pathology & 73.2(72.4,73.0) & 81.8(81.8,82.7) &15,496 & 8.6 &$4.58\times10^{-3}$\\
					Ultrasound & 93.8(92.3,94.4) & 98.4(96.9,99.3) &10,193 & 4.6 &$1.09\times10^{-3}$\\
					Fundus & 92.4(92.3,92.9) & 98.5(97.3,99.5) &805 & 6.1 &$3.52\times10^{-6}$\\
					Dermoscopy & 51.6(50.4,51.6) & 90.8(90.3,91.9) &65 & 39.2 &$4.90\times10^{-7}$\\
					Endoscope & 93.0(92.5,93.6) & 95.5(94.9,96.2) &410 & 2.5 &$1.14\times10^{-2}$\\
					OCT & 98.0(97.8,98.1) & 98.0(97.3,98.1) &283 & 0.0 &$4.03\times10^{-1}$\\
					PET & 86.8(85.9,88.4) & 92.5(91.8,93.2) &1,561 & 5.7 &$3.35\times10^{-3}$\\
					\bottomrule
				\end{tabular}
			}
		\end{table*}
		
		\clearpage
		\newpage
		
		\begin{figure*}[!h]
			\centering
			\includegraphics[width=0.75\linewidth]{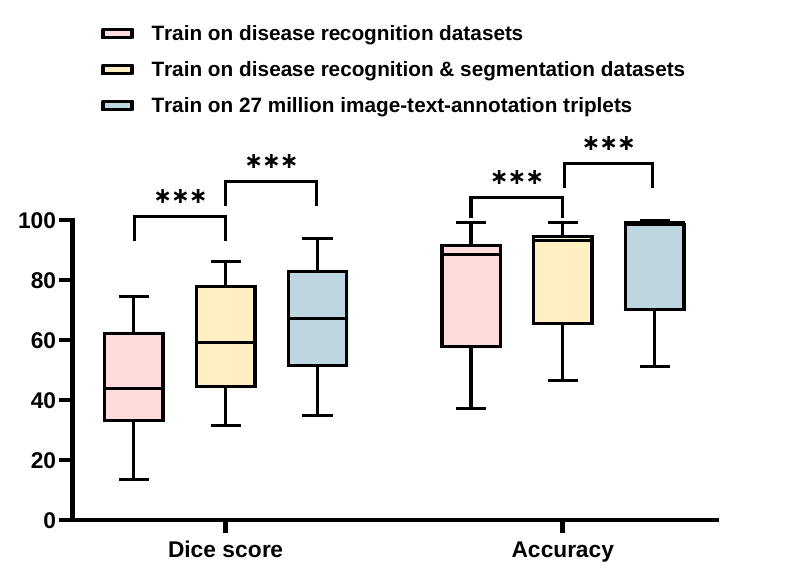}
			\caption{\textbf{Ablation studies of grounded disease recognition}. We conduct ablation studies to demonstrate that universal training on different types of datasets can effectively improve the Dice scores and accuracy of grounded disease recognition. We evaluate the effectiveness of training datasets: (1) only train on disease recognition datasets (\textbf{Extended Data Table~\ref{table:dataset_grounded_disease}}); (2) add segmentation datasets for segmentation training; (3) train on the whole 27 million triplet dataset. Specifically, segmentation training improves dice scores by $20.32\%$ (***$P<1\times10^{-3}$) and accuracy by $4.90\%$ (***$P<1\times10^{-3}$), respectively. Universal training on 27 million triplets improves dice scores by $7.15\%$ (***$P<1\times10^{-3}$) and accuracy by $4.37\%$ (***$P<1\times10^{-3}$), respectively. These findings robustly validate the effectiveness of universal training in improving grounded biomedical image interpretation.
			}
			\label{fig_extended_ablation}
		\end{figure*}
		
		\clearpage

		\begin{figure*}[!h]
			\centering
			\includegraphics[width=0.85\linewidth]{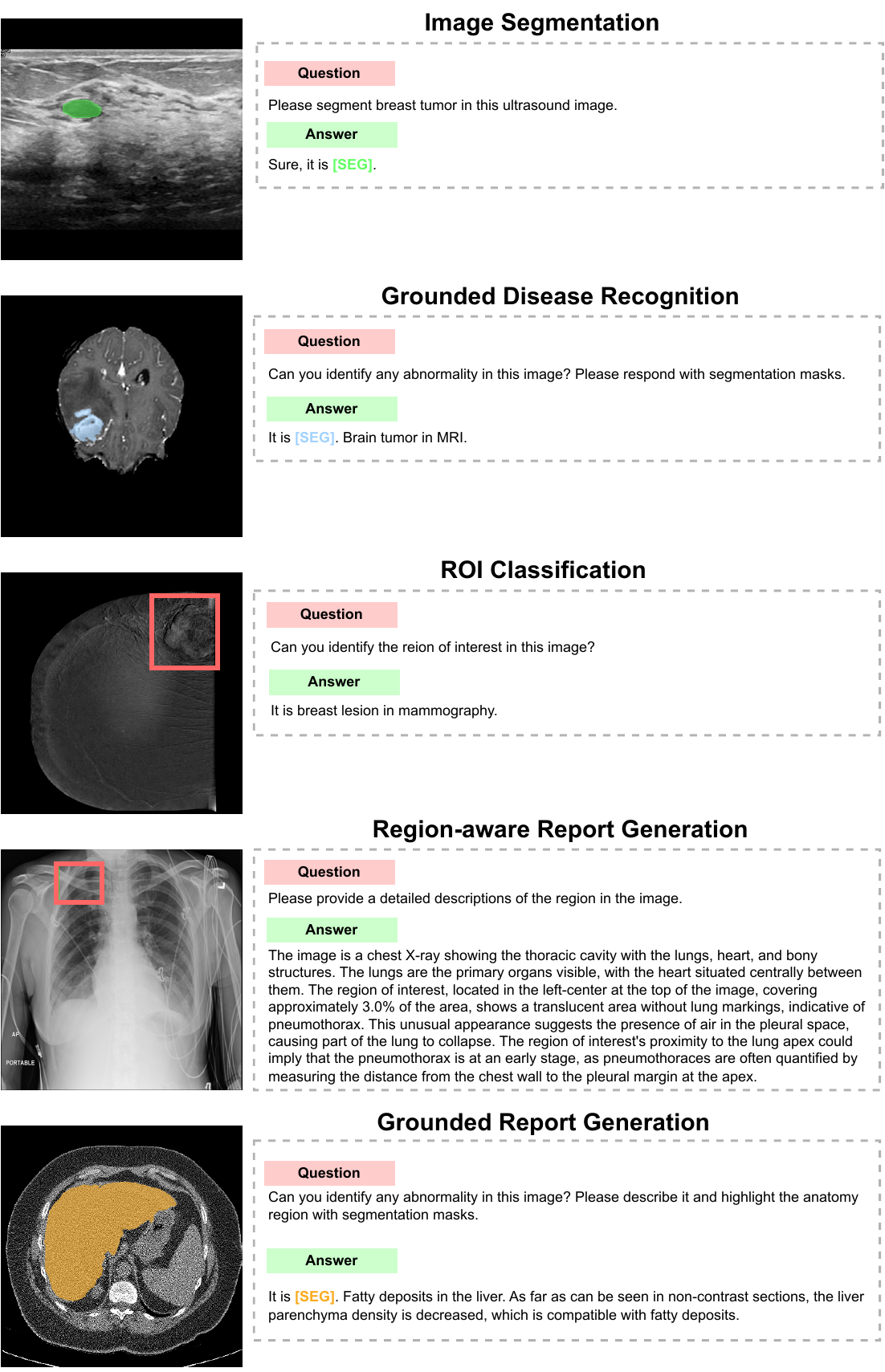}
			\caption{Uniform VQA format for universal training. For different biomedical tasks, we process the datasets into the same format as above. We present the data format for image segmentation, disease recognition, ROI classification, region-aware report generation, and grounded report generation.}
			\label{fig_extended_vqa_format}
		\end{figure*}
		
		\clearpage

		\begin{figure*}[!h]
			\centering
			\includegraphics[width=1\linewidth]{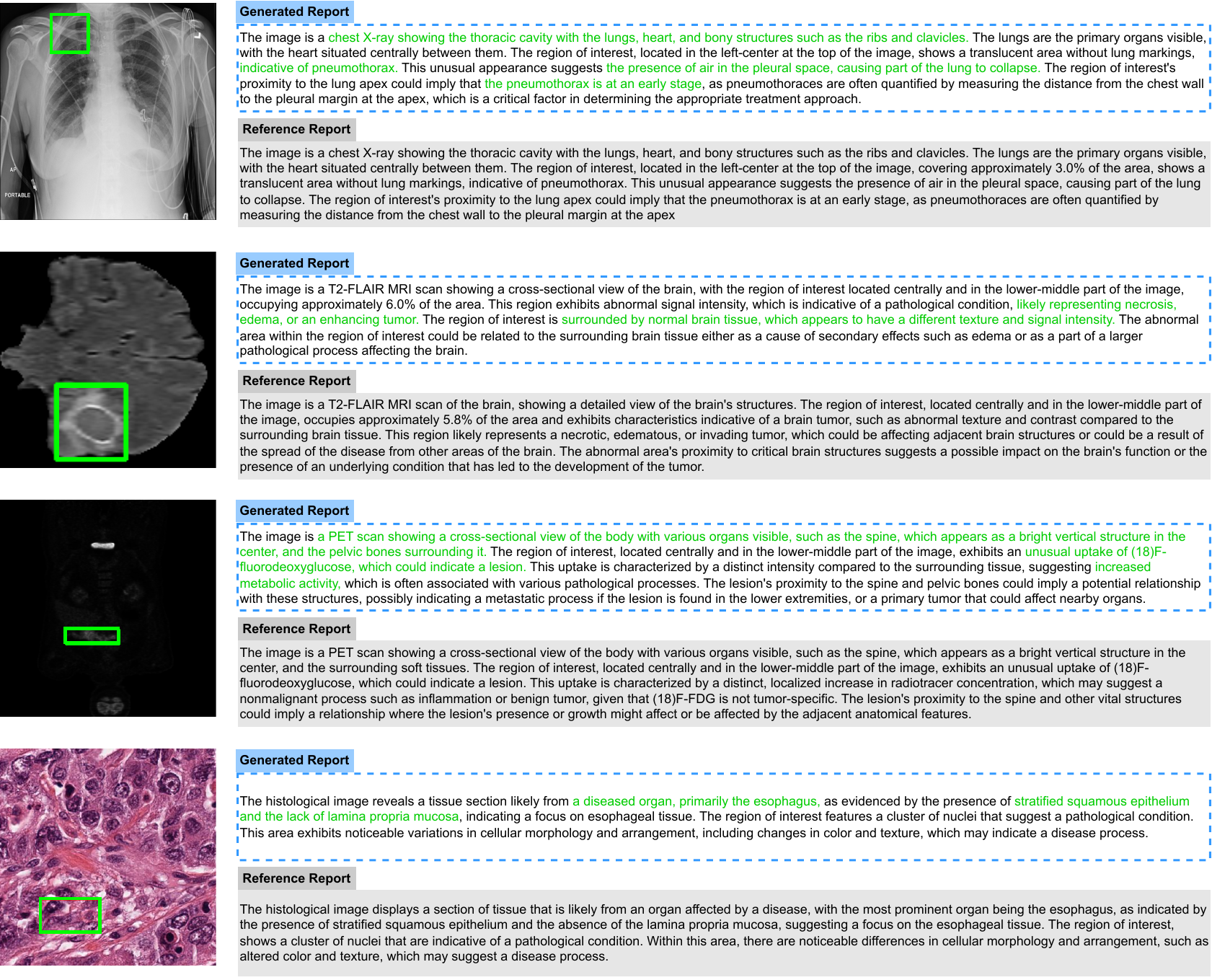}
			\caption{Region-aware report generation results on the MedTrinity~\cite{medtrinity} dataset. The text in \textcolor{mygreen}{green} indicates the correct contents in generated reports.}
			\label{fig_extended_roi_report}
		\end{figure*}
		
		\clearpage
		
		\begin{figure*}[!h]
			\centering
			\includegraphics[width=1\linewidth]{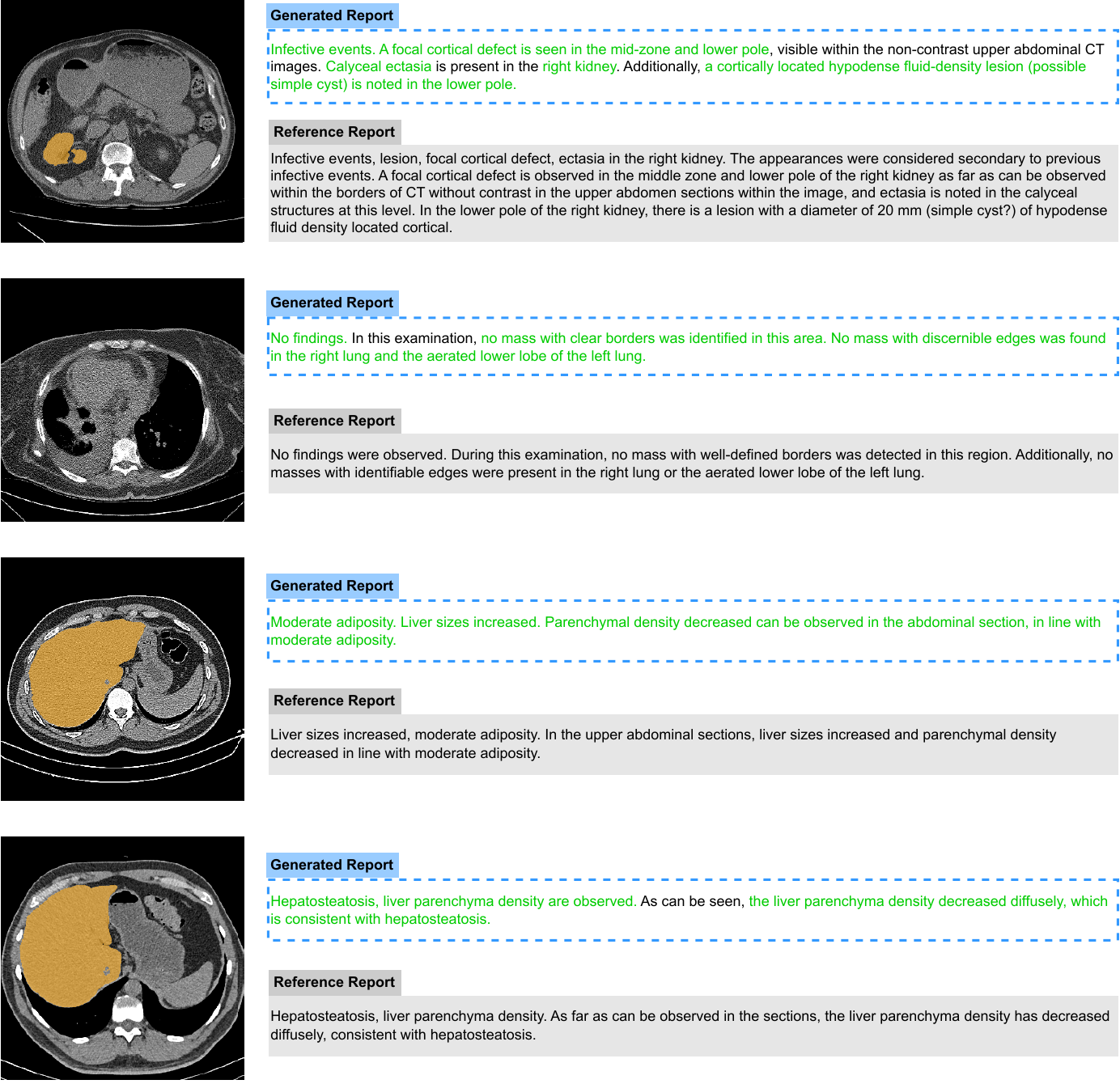}
			\caption{Grounded report generation results on the RadGenome~\cite{radgenome} dataset. The generated segmentation masks are in \textcolor{myorange}{orange}, which indicate the location of the corresponding organs described in the reports. The text in \textcolor{mygreen}{green} indicates the correct contents in generated reports.}
			\label{fig_extended_grg}
		\end{figure*}
		
		\clearpage

		\begin{table*}
			\centering
			\caption{The public codes of methods used in this study.
			}
			\label{table_code_source}
			\begin{tabular}{ll}
				\toprule[1.2pt]
				\textbf{Method} &\textbf{Sources}\\
				\hline
				
				SAM2~\cite{sam2} &\url{https://github.com/facebookresearch/sam2}\\
				InternVL~\cite{internvl} & \url{https://github.com/OpenGVLab/InternVL}\\
				
				SA2VA~\cite{sa2va} & \url{https://github.com/magic-research/Sa2VA}\\
				
				LISA~\cite{lisa} & \url{https://github.com/dvlab-research/LISA}\\
				GLaMM~\cite{glamm} & \url{https://github.com/mbzuai-oryx/groundingLMM}\\
				
				Osprey~\cite{osprey} & \url{https://github.com/CircleRadon/Osprey}\\
				
				VoCo~\cite{voco,voco-v1} & \url{https://github.com/Luffy03/Large-Scale-Medical}\\
				
				
				MedSAM~\cite{medsam} & \url{https://github.com/bowang-lab/MedSAM}\\
				
				BiomedParse~\cite{biomedparse} & \url{https://github.com/microsoft/BiomedParse}\\
				
				SegVol~\cite{segvol} & \url{https://github.com/BAAI-DCAI/SegVol}\\
				
				SAT~\cite{SAT} & \url{https://github.com/zhaoziheng/SAT}\\
				
				LLaVA-Med~\cite{Llava-med} & \url{https://github.com/microsoft/LLaVA-Med}\\
				
				MedRegA~\cite{MedRegA} &\url{https://github.com/xmed-lab/MedRegA}\\
				MedPLIB~\cite{MedPLIB} &\url{https://github.com/ShawnHuang497/MedPLIB}\\
				MedTrinity~\cite{medtrinity} &\url{https://github.com/UCSC-VLAA/MedTrinity-25M}\\
				\toprule[1.2pt]
			\end{tabular}
			
		\end{table*}
		
		\clearpage
		
		\begin{table*}
			\centering
			\caption{\textbf{Architecture details and Training settings.} 
			}
			\label{table_preprocess}
			\begin{tabular}{l|l}
				\toprule[1.2pt]
				MLLM &InternVL2.5 \\
				Segmentation Model &SAM2-hiera-large \\
				Network Params &1.23 B\\
				Segmentation Loss &Dice+Cross Entropy\\
				LLM Loss &Cross Entropy\\
				\hline
				
				Platform &Pytorch 2.3.1\\
				CUDA version &11.8\\
				Training GPU(s) &8*NVIDIA H800 (80G)\\
				Inference GPU(s) &1*NVIDIA 3090 (24G)\\
				Training epochs &10\\
				Training time &120 hours\\
				Batch size &32\\
				Optimizer &AdamW\\
				Scheduler &Warmup+Cosine\\
				Learning rate &4e-5\\
				Image size &$1024 \times 1024$\\
				Inference time &Average 0.175 s/image\\
				\toprule[1.2pt]
			\end{tabular}
		\end{table*}
		
	\end{appendices}
\end{document}